\documentclass[preprint,12pt]{elsarticle}

\usepackage{amssymb}
\usepackage{amsmath}
\usepackage{xcolor}
\usepackage{tikz}
\usetikzlibrary{arrows.meta, positioning, fit, backgrounds}
\usepackage{booktabs}
\usepackage{xltabular}
\usepackage{array}
\usepackage{ragged2e}

\newcolumntype{Y}{>{\RaggedRight\arraybackslash}X}
\journal{Computer Science Review}

\begin{document}

\begin{frontmatter}

%% Title, authors and addresses

%% use the tnoteref command within \title for footnotes;
%% use the tnotetext command for theassociated footnote;
%% use the fnref command within \author or \address for footnotes;
%% use the fntext command for theassociated footnote;
%% use the corref command within \author for corresponding author footnotes;
%% use the cortext command for theassociated footnote;
%% use the ead command for the email address,
%% and the form \ead[url] for the home page:
%% \title{Title\tnoteref{label1}}
%% \tnotetext[label1]{}
%% \author{Name\corref{cor1}\fnref{label2}}
%% \ead{email address}
%% \ead[url]{home page}
%% \fntext[label2]{}
%% \cortext[cor1]{}
%% \affiliation{organization={},
%%             addressline={},
%%             city={},
%%             postcode={},
%%             state={},
%%             country={}}
%% \fntext[label3]{}

\title{From Pixels to Portraits: A Comprehensive Survey of Talking Head Generation Techniques and Applications}

%% use optional labels to link authors explicitly to addresses:
%% \author[label1,label2]{}
%% \affiliation[label1]{organization={},
%%             addressline={},
%%             city={},
%%             postcode={},
%%             state={},
%%             country={}}
%%
%% \affiliation[label2]{organization={},
%%             addressline={},
%%             city={},
%%             postcode={},
%%             state={},
%%             country={}}

\author[inst1]{Shreyank N Gowda}

\affiliation[inst1]{organization={School of Computer Science, University of Nottingham},%Department and Organization 
            city={Nottingham},
            country={United Kingdom}}
\affiliation[inst2]{organization={Independent Researcher}}

\author[inst2]{Dheeraj Pandey}
\author[inst2]{Shashank Narayana Gowda}

\begin{abstract}
Talking head generation has progressed rapidly from landmark- and GAN-based facial animation to diffusion models, neural rendering, 3D-aware avatars, and foundation-model-assisted systems. This progress has enabled increasingly realistic audio-, image-, and video-driven talking heads, but it has also made the field difficult to navigate because methods differ substantially in their inputs, assumptions, controllability, temporal stability, computational cost, and risks of misuse. This survey provides a critical review of talking head generation techniques, organizing the literature into four broad families: image-driven, audio-driven, video-driven, and 3D/neural-rendering-based approaches. For each family, we discuss the underlying technical ideas, representative methods, strengths, limitations, datasets, and evaluation practices. Beyond cataloguing prior work, we analyse the persistent gap between commonly reported quantitative metrics and perceptual quality, and compare publicly available models in terms of inference time, memory requirements, and human-rated visual quality. We also examine emerging trends, including diffusion-based generation, 3D-aware representation learning, controllable emotional expression, real-time deployment, and the growing importance of provenance, watermarking, and deepfake detection. Finally, we identify open challenges around robust evaluation, identity preservation, lip synchronisation, temporal consistency, demographic fairness, computational efficiency, and responsible use. This review aims to provide researchers and practitioners with a structured and up-to-date map of the talking head generation landscape, while highlighting the technical and societal questions that should shape future work.
\end{abstract}

%%Graphical abstract
%\begin{graphicalabstract}
%\includegraphics{grabs}
%\end{graphicalabstract}

%%Research highlights
%\begin{highlights}
%\item Research highlight 1
%\item Research highlight 2
%\end{highlights}

\begin{keyword}
%% keywords here, in the form: keyword \sep keyword
Talking head \sep facial animation \sep lip movement \sep facial expression
%% PACS codes here, in the form: \PACS code \sep code
%\PACS 0000 \sep 1111
%% MSC codes here, in the form: \MSC code \sep code
%% or \MSC[2008] code \sep code (2000 is the default)
%\MSC 0000 \sep 1111
\end{keyword}

\end{frontmatter}

%% \linenumbers

%% main text
\section{Introduction}
\label{sec:intro}

Talking head generation aims to synthesize realistic videos of a person speaking, typically from a still image, a reference video, an audio signal, text, or a combination of these inputs. The task sits at the intersection of computer vision, graphics, speech processing, multimodal learning, and human-computer interaction. A convincing talking head must preserve identity, synchronize lip movements with speech, model facial expression and head motion, remain temporally stable, and generate visually plausible details under changes in pose, emotion, language, and speaking style. These requirements make talking head generation a challenging problem, but also a practically important one, with applications in digital avatars, video conferencing, film and game production, virtual reality, assistive communication, education, and embodied conversational agents.

Early work in this area relied on rule-based systems and explicit mappings from speech or text to lip motion~\cite{bregler1997video,xie2007realistic}. These methods demonstrated the promise of speech-driven facial animation, but were limited in visual realism, expression diversity, and generalization to unconstrained identities and environments. The widespread adoption of deep learning changed the trajectory of the field. Convolutional networks, recurrent models, generative adversarial networks, attention mechanisms, transformers, and more recently diffusion and neural rendering models have enabled data-driven systems that can synthesize increasingly natural facial motion and high-quality portrait videos~\cite{resnet,densenet,gans,vaswani2017attention,tpsm,zhang2023sadtalker,dagan,daganplus,doukas2021headgan,wang2021audio2head,gowda2024fe}. As large-scale face, speech, and video datasets became available, talking head generation moved from controlled lip animation towards more general portrait animation, including identity-preserving reenactment, audio-driven facial dynamics, emotion-aware synthesis, 3D-aware avatars, and real-time interactive generation.

The field has also evolved technically. Earlier neural methods often relied on landmarks, 3D morphable models, optical flow, or keypoint-based motion transfer to separate identity from motion. GAN-based approaches improved photorealism and image-level fidelity, while attention mechanisms helped models focus on semantically important regions such as the mouth, eyes, and facial contours~\cite{mirza2014conditional,was,rhm,dagan}. Transformer-based and codebook-based methods later improved long-range temporal modeling and controllability. More recently, diffusion-based approaches have become increasingly prominent because of their ability to model complex audio-visual distributions and generate expressive, high-quality portrait videos under weak conditioning~\cite{emo2024,hallo2024,hallo22024}. In parallel, neural radiance fields, 3D-aware representations, and Gaussian splatting-based avatars have shifted part of the field towards explicit or implicit 3D representations, enabling improved view consistency, controllability, and real-time rendering~\cite{gaussianavatars2024,splattingavatar2024}. These developments have expanded the scope of talking head generation from facial animation alone to the broader problem of controllable, photorealistic, and interactive human avatar synthesis.

Despite this progress, the field remains difficult to assess. Talking head generation methods differ substantially in their inputs, assumptions, training data, target applications, and evaluation protocols. Some methods animate a single image using a driving video, some generate facial motion directly from audio, some edit an existing video, while others construct a controllable 3D avatar. These systems are often evaluated on different datasets, under different levels of supervision, and with different preprocessing choices. As a result, reported quantitative results are not always directly comparable. Moreover, commonly used metrics such as PSNR and SSIM primarily measure pixel-level similarity and often fail to capture the perceptual realism of talking heads, including lip synchronization, expression naturalness, identity preservation, temporal coherence, and the absence of distracting artifacts~\cite{hore2010image}. Metrics such as FID, LPIPS, identity similarity, landmark distance, and lip-sync confidence provide complementary signals, but no single metric reliably captures the overall quality of a generated talking head. Human evaluation remains important, but it is costly, subjective, and difficult to reproduce.

This evaluation gap is particularly important because talking head generation is no longer only a laboratory problem. Modern systems are increasingly close to real-world deployment in telepresence, virtual assistants, entertainment, online education, and social media content creation. Practical deployment requires more than high visual quality: models must be efficient, controllable, robust to diverse identities and languages, and safe to use. Real-time inference, GPU memory usage, latency, and stability over long videos are therefore central concerns. At the same time, the same technology can be misused for impersonation, misinformation, and non-consensual synthetic media. Any contemporary review of talking head generation must therefore consider not only model architectures and benchmark scores, but also provenance, watermarking, deepfake detection, consent, dataset bias, demographic fairness, and responsible deployment.

This survey provides a structured and critical review of talking head generation. We organize the literature into four broad families: image-driven, audio-driven, video-driven, and 3D/neural-rendering-based approaches. Image-driven methods animate a source portrait using visual cues such as landmarks, keypoints, pose, or expression from another image or video. Audio-driven methods synthesize lip motion, facial expressions, and head movements from speech signals. Video-driven methods focus on reenactment, retalking, and motion transfer from a reference sequence. 3D and neural-rendering methods, including NeRF-based and Gaussian splatting-based avatars, aim to improve view consistency, controllability, and photorealistic rendering. For each family, we discuss representative methods, technical assumptions, strengths, limitations, and typical failure cases.

Beyond cataloguing prior work, this survey places particular emphasis on evaluation and practical usability. We review commonly used datasets and metrics, discuss why existing evaluation protocols often disagree with perceptual quality, and compare publicly available models in terms of inference time, memory requirements, and human-rated visual quality. This comparison is intended to complement standard benchmark numbers by highlighting the trade-offs that matter in real applications, where a method must not only look realistic but also run efficiently and consistently. We also identify emerging trends, including diffusion-based portrait animation, long-duration generation, emotion and style control, real-time audio-driven avatars, 3D-aware synthesis, and responsible generation.

In summary, this survey makes the following contributions:
\begin{itemize}
    \item We provide an updated taxonomy of talking head generation methods, covering image-driven, audio-driven, video-driven, and 3D/neural-rendering-based approaches.
    \item We review the evolution of the field from rule-based and landmark-driven systems to GANs, transformers, diffusion models, NeRF-based methods, and Gaussian avatar representations.
    \item We analyse datasets and evaluation metrics, highlighting the mismatch between commonly reported quantitative scores and perceived visual quality.
    \item We compare publicly available models using practical criteria such as inference time, GPU memory usage, and human-rated quality.
    \item We discuss open challenges around lip synchronization, identity preservation, emotional expressiveness, temporal consistency, demographic robustness, computational efficiency, provenance, and responsible use.
\end{itemize}

By synthesizing technical progress, evaluation practice, deployment constraints, and societal risks, this survey aims to provide researchers and practitioners with an up-to-date map of the talking head generation landscape and a clear view of the challenges that should shape future work.

\section{What Comprises a Good Talking Head Generation?}
\label{sec:what}

A talking head generation system is successful only when it satisfies several constraints simultaneously. A generated video may be visually sharp but poorly synchronized with speech; it may achieve accurate lip motion but fail to preserve identity; or it may perform well on short benchmark clips while breaking under long-duration generation, large head poses, emotional speech, multilingual audio, or real-time deployment. Therefore, the quality of a talking head should not be reduced to a single visual metric. Instead, it should be understood as a multi-objective problem involving identity, synchronization, visual realism, temporal stability, controllability, efficiency, and responsible use.

Prior work~\cite{what} formalized several desirable properties for synthesized talking-head videos, including identity preservation, lip synchronization, visual quality, and spontaneous natural motion. These properties remain central. However, the field has moved substantially beyond the setting considered by early benchmarks. Recent methods now include efficient implicit-keypoint systems~\cite{liveportrait2024}, real-time audio-driven models~\cite{vasa2024,teller2025,read2025}, diffusion and video-transformer-based portrait animation~\cite{dimitra2025,hallo32025,omnihuman2025}, one-shot and large-scale 3D avatar models~\cite{real3dportrait2024,lam2025}, and Gaussian-splatting-based talking head avatars~\cite{gaussianspeech2025,pgstalker2025,unigaha2025,vasa3d2025,gaussianemotalker2026}. As a result, a contemporary definition of a ``good'' talking head must also include long-range consistency, emotional expressiveness, controllability, view consistency, latency, and safety.

We summarize the main quality dimensions as follows:

\begin{itemize}
    \item \textbf{Identity preservation:} The generated video should maintain the source subject's identity across frames, expressions, poses, and lighting conditions. This is commonly evaluated using face-recognition embeddings, for example by computing cosine similarity between the source and generated identities.

    \item \textbf{Lip synchronization:} The mouth motion should match the spoken content at both phonetic and semantic levels. Good lip-sync requires not only plausible mouth opening and closing, but also correct viseme formation, stable teeth and tongue appearance, and robustness across speakers, accents, languages, and speaking rates.

    \item \textbf{Visual fidelity:} Individual frames should be realistic, sharp, and free from local artifacts around the mouth, teeth, eyes, hair, face boundary, and background. Metrics such as PSNR, SSIM, FID, LPIPS, and CPBD are often used, but they only partially capture perceived realism~\cite{hore2010image}.

    \item \textbf{Temporal consistency:} The generated video should remain stable over time. Flickering, identity drift, unstable mouth interiors, inconsistent eye gaze, and frame-to-frame texture changes can make even high-quality individual frames appear unrealistic in video form.

    \item \textbf{Natural motion and expressiveness:} A convincing talking head should include natural head motion, blinking, gaze changes, facial expressions, and emotion-consistent dynamics. Recent works increasingly model not only lip motion but also full facial dynamics, head pose, emotional intensity, and upper-body or accessory motion~\cite{vasa2024,dimitra2025,teller2025,gaussianemotalker2026}.

    \item \textbf{Controllability:} Practical systems should allow control over relevant attributes such as expression, emotion, gaze, pose, speaking style, viewpoint, and sometimes body motion or background. This is especially important for avatars, editing tools, and professional content creation~\cite{liveportrait2024,edityourself2026,omnihuman2025}.

    \item \textbf{View and geometry consistency:} 3D-aware and Gaussian-splatting-based methods introduce an additional requirement: the generated head should remain consistent under camera movement and novel views. This is important for virtual reality, telepresence, and interactive avatars~\cite{real3dportrait2024,gaussianspeech2025,unigaha2025,vasa3d2025}.

    \item \textbf{Generalization and robustness:} A strong model should generalize across identities, ages, ethnicities, genders, lighting conditions, head poses, occlusions, accents, emotions, and languages. Robustness is particularly important because many commonly used datasets contain biases in speaker demographics, recording conditions, and video quality.

    \item \textbf{Efficiency and deployability:} Real-world systems require acceptable inference time, GPU memory usage, latency, and model size. For video conferencing, virtual assistants, gaming, and telepresence, a model that is visually impressive but slow or expensive to run may be unsuitable for deployment~\cite{vasa2024,liveportrait2024,read2025,teller2025,pgstalker2025}.

    \item \textbf{Safety and responsible use:} Talking head generation can be misused for impersonation, misinformation, and non-consensual synthetic media. Responsible deployment therefore requires attention to consent, provenance, watermarking, dataset governance, misuse prevention, and compatibility with deepfake detection systems.
\end{itemize}

These criteria are often in tension. Improving expressiveness may increase the risk of identity drift; increasing resolution may reduce real-time performance; optimizing for lip synchronization may produce unnatural head motion; and enforcing temporal smoothness may suppress spontaneous expression. The relative importance of each criterion therefore depends on the application. A video conferencing system may prioritize latency, identity preservation, and stability, while a film-production tool may prioritize high-resolution controllability and editability. An assistive communication system may require intelligible lip motion and emotional clarity, while a virtual reality avatar may require real-time performance and view-consistent 3D rendering.

\subsection{Fundamentals in Talking Head Generation}

The central goal of talking head generation is to synthesize a video of a target identity speaking or moving according to a driving signal. The target identity may be provided as a still image, a short video, a learned avatar, a 3D representation, or a Gaussian-splatting-based head model. The driving signal may be audio, text, another video, facial landmarks, keypoints, 3D morphable model parameters, expression codes, pose trajectories, diffusion latents, or combinations of these. In a general form, the task can be written as:
\begin{equation}
    V_g = f(I_s, c; \theta),
\end{equation}
where $I_s$ denotes the source identity or appearance reference, $c$ denotes the conditioning signal, $V_g$ is the generated talking-head video, and $f$ is a generative model with parameters $\theta$. In video-driven methods, $c$ may be a driving video $V_d$; in audio-driven methods, it may be an audio signal $A$; in text-driven or editing methods, it may include text or transcript information; and in 3D-aware methods, it may include camera, pose, geometry, or expression parameters.

Although architectures differ, many talking head generation systems can be understood through four broad stages: representation, motion or condition modeling, synthesis, and refinement.

\paragraph{Representation.}
The model first extracts representations of identity and conditioning information:
\begin{equation}
    z_s = E_s(I_s; \theta_s), \quad z_c = E_c(c; \theta_c),
\end{equation}
where $E_s$ encodes source identity or appearance, and $E_c$ encodes the driving signal. Depending on the method, $z_c$ may represent landmarks, keypoints, optical flow, audio embeddings, phoneme features, expression coefficients, head pose, 3D geometry, diffusion conditioning tokens, or neural-rendering features. Earlier approaches often relied on landmarks, 3DMM coefficients, or keypoints~\cite{siarohin2019first,wang2021one}, while recent approaches increasingly use richer latent representations learned by diffusion models, transformers, or 3D Gaussian avatar models~\cite{vasa2024,hallo32025,read2025,gaussianspeech2025,vasa3d2025}.

\paragraph{Motion and condition modeling.}
The model then maps the conditioning representation to a motion, expression, or control representation:
\begin{equation}
    m = M(z_s, z_c; \theta_m),
\end{equation}
where $m$ may denote dense motion fields, facial landmarks, 3DMM coefficients, head pose, expression codes, attention maps, diffusion latents, deformation fields, or Gaussian offsets. This stage is central because it determines how speech, pose, and expression are transferred while preserving identity. For example, implicit-keypoint methods learn compact motion representations for efficient animation~\cite{liveportrait2024}, diffusion-transformer methods model audio-to-motion or audio-to-video distributions~\cite{dimitra2025,read2025}, and Gaussian avatar methods predict audio-driven deformations in a 3D representation~\cite{gaussianspeech2025,pgstalker2025,unigaha2025}.

\paragraph{Synthesis.}
A generator, decoder, renderer, or denoising model produces the output frames:
\begin{equation}
    V_g = G(z_s, m; \theta_g).
\end{equation}
In 2D methods, $G$ is often an image or video generator. In GAN-based systems, $G$ is trained adversarially to produce photorealistic frames~\cite{goodfellow2014generative}. In diffusion-based systems, synthesis may proceed through iterative denoising in pixel space or latent space~\cite{hallo32025,dimitra2025,read2025}. In 3D-aware methods, $G$ may instead be a neural renderer using a mesh, triplane, NeRF, or Gaussian representation~\cite{real3dportrait2024,lam2025,gaussianspeech2025,vasa3d2025}.

\paragraph{Refinement.}
Many systems include refinement stages to improve mouth details, teeth stability, eye motion, high-frequency texture, background consistency, or temporal smoothness. These refinements may involve local mouth generators, face enhancement, super-resolution, temporal discriminators, optical-flow constraints, video-to-video editing, or region-aware inpainting~\cite{edityourself2026}. Such components are particularly important for high-resolution output and long-duration generation.

\paragraph{Training objectives.}
Talking head generation models are usually trained with a combination of losses. A reconstruction loss encourages the generated video to match the target video:
\begin{equation}
    L_{\text{rec}} = \lVert V - V_g \rVert_1.
\end{equation}

A perceptual loss compares high-level visual features extracted by a pre-trained network:
\begin{equation}
    L_{\text{perc}} = \lVert \phi(V) - \phi(V_g) \rVert_2,
\end{equation}
where $\phi(\cdot)$ denotes a feature extractor such as VGG~\cite{simonyan2014very}.

An adversarial loss encourages photorealistic outputs:
\begin{equation}
    L_{\text{adv}} =
    \mathbb{E}_{V \sim p_{\text{data}}}[\log D_{\text{adv}}(V)] +
    \mathbb{E}_{V_g \sim p_{\text{gen}}}[\log(1 - D_{\text{adv}}(V_g))],
\end{equation}
where $D_{\text{adv}}$ is a discriminator.

Identity preservation can be encouraged using a face-recognition embedding:
\begin{equation}
    L_{\text{id}} = 1 - \cos(\psi(I_s), \psi(V_g)),
\end{equation}
where $\psi(\cdot)$ is an identity encoder.

Lip synchronization can be encouraged by comparing audio and visual speech embeddings:
\begin{equation}
    L_{\text{sync}} = d(\eta_a(A), \eta_v(V_g)),
\end{equation}
where $\eta_a(\cdot)$ and $\eta_v(\cdot)$ encode audio and visual speech features, and $d(\cdot)$ is a distance function.

Temporal stability can be encouraged through frame-difference, optical-flow, or feature-consistency losses:
\begin{equation}
    L_{\text{temp}} = \sum_t \lVert \Delta V_t - \Delta V_{g,t} \rVert_1.
\end{equation}

The full training objective is typically a weighted combination of complementary terms:
\begin{equation}
    L_{\text{total}} =
    \lambda_{\text{rec}}L_{\text{rec}} +
    \lambda_{\text{perc}}L_{\text{perc}} +
    \lambda_{\text{adv}}L_{\text{adv}} +
    \lambda_{\text{id}}L_{\text{id}} +
    \lambda_{\text{sync}}L_{\text{sync}} +
    \lambda_{\text{temp}}L_{\text{temp}}.
\end{equation}

Not all methods use all these objectives. Keypoint-based reenactment methods often emphasize reconstruction, perceptual, equivariance, and motion losses~\cite{siarohin2019first,wang2021one,liveportrait2024}. Audio-driven methods often include synchronization and expression-related losses~\cite{wang2021audio2head,zhang2023sadtalker,vasa2024,dimitra2025}. Diffusion-based methods optimize denoising or score-matching objectives, conditioned on audio, pose, identity, or reference frames~\cite{hallo32025,read2025,edityourself2026}. Neural-rendering and Gaussian-splatting methods may include geometry, deformation, photometric, landmark, perceptual, wrinkle, and view-consistency losses~\cite{gaussianspeech2025,unigaha2025,vasa3d2025,gaussianemotalker2026}.

\begin{figure}
    \centering
    \includegraphics[width=0.95\linewidth]{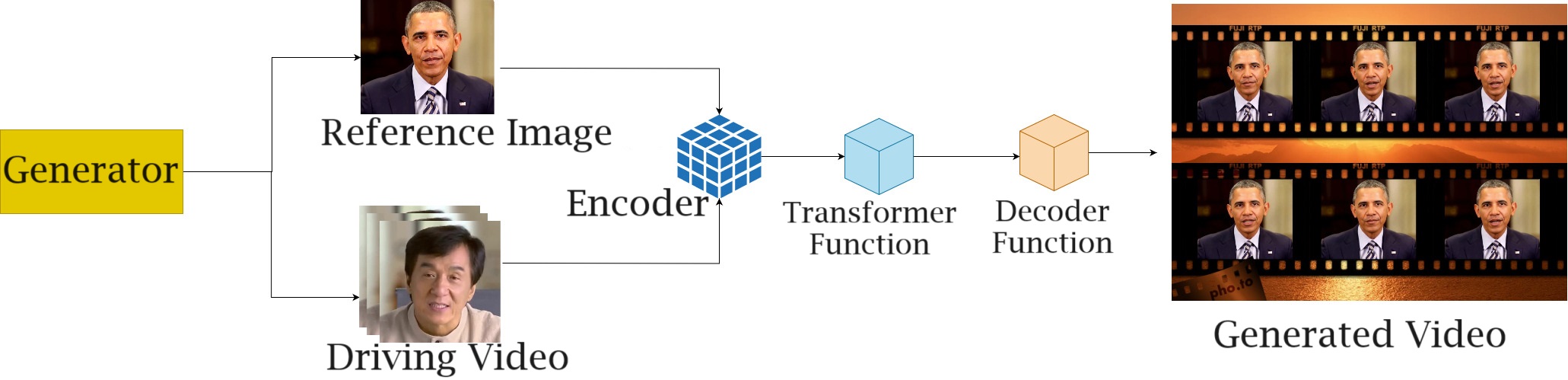}
    \caption{A generic talking head generation pipeline. The source identity is combined with a driving signal such as video, audio, text, landmarks, pose, expression coefficients, diffusion latents, or 3D parameters. Different model families instantiate this pipeline differently, using 2D generators, diffusion models, neural renderers, or 3D/Gaussian avatar representations.}
    \label{fig:overview}
\end{figure}

Figure~\ref{fig:overview} illustrates a representative pipeline, but modern talking head generation methods differ substantially in how they encode identity, represent motion, impose temporal consistency, and synthesize output frames. This diversity motivates the taxonomy used in the remainder of this survey.

\section{Taxonomy of Talking Head Generation Methods}
\label{sec:classes}

Talking head generation methods can be categorized in several ways: by the driving signal, by the internal representation, by the generative backbone, or by the intended application. Earlier surveys often grouped methods mainly according to the input modality, such as image-driven, audio-driven, video-driven, or 3D-based generation. This remains useful, but the field has evolved substantially. Recent methods increasingly combine multiple conditioning signals, use diffusion or video-transformer backbones, rely on 3D-aware or Gaussian-splatting representations, and target real-time or interactive deployment. Therefore, we adopt a broader taxonomy that reflects both the driving modality and the representation/generation paradigm.

Figure~\ref{fig:new-taxonomy} summarizes the taxonomy used in this survey. The categories are not mutually exclusive. For example, VASA-1 is audio-driven but also designed for real-time generation~\cite{vasa2024}; EditYourself is both diffusion-based and editing-oriented~\cite{edityourself2026}; and GaussianSpeech is both audio-driven and 3D Gaussian-splatting-based~\cite{gaussianspeech2025}. The taxonomy should therefore be read as a way of understanding the dominant design choice of each family, rather than as a strict partition.

\begin{figure}
\centering
\resizebox{\textwidth}{!}{
\begin{tikzpicture}[
    font=\small,
    root/.style={
        rectangle,
        rounded corners,
        draw=black,
        thick,
        align=center,
        minimum width=4.8cm,
        minimum height=0.9cm,
        fill=gray!15
    },
    family/.style={
        rectangle,
        rounded corners,
        draw=black,
        thick,
        align=center,
        text width=3.45cm,
        minimum height=1.05cm,
        fill=blue!8
    },
    sub/.style={
        rectangle,
        rounded corners,
        draw=black!70,
        align=center,
        text width=3.25cm,
        minimum height=0.72cm,
        fill=white
    },
    cross/.style={
        rectangle,
        rounded corners,
        draw=black!70,
        dashed,
        align=center,
        text width=17.6cm,
        minimum height=0.8cm,
        fill=orange!10
    },
    arrow/.style={-Latex, thick}
]

\node[root] (root) at (0,0) {Talking Head Generation};

\node[family] (vis) at (-7.2,-2.0) {2D visual/video-driven animation};
\node[family] (aud) at (-3.6,-2.0) {Audio-driven talking heads};
\node[family] (diff) at (0,-2.0) {Diffusion and foundation video models};
\node[family] (threeD) at (3.6,-2.0) {3D, NeRF, and Gaussian avatars};
\node[family] (edit) at (7.2,-2.0) {Text, semantic, and editing control};

\draw[arrow] (root) -- (vis);
\draw[arrow] (root) -- (aud);
\draw[arrow] (root) -- (diff);
\draw[arrow] (root) -- (threeD);
\draw[arrow] (root) -- (edit);

\node[sub] (vis1) at (-7.2,-3.35) {Face reenactment};
\node[sub] (vis2) at (-7.2,-4.25) {Motion retargeting};
\node[sub] (vis3) at (-7.2,-5.15) {Expression and attribute transfer};
\node[sub] (vis4) at (-7.2,-6.05) {High-resolution and efficient animation};

\node[sub] (aud1) at (-3.6,-3.35) {Lip synchronization};
\node[sub] (aud2) at (-3.6,-4.25) {Audio-to-face and head motion};
\node[sub] (aud3) at (-3.6,-5.15) {Emotion, style, and prosody};
\node[sub] (aud4) at (-3.6,-6.05) {Real-time and streaming generation};

\node[sub] (diff1) at (0,-3.35) {Audio-to-video diffusion};
\node[sub] (diff2) at (0,-4.25) {Video diffusion transformers};
\node[sub] (diff3) at (0,-5.15) {Large-scale human animation};
\node[sub] (diff4) at (0,-6.05) {Long-duration generation};

\node[sub] (d1) at (3.6,-3.35) {3DMM and mesh avatars};
\node[sub] (d2) at (3.6,-4.25) {NeRF and implicit fields};
\node[sub] (d3) at (3.6,-5.15) {3D Gaussian splatting};
\node[sub] (d4) at (3.6,-6.05) {View-consistent rendering};

\node[sub] (e1) at (7.2,-3.35) {Text-driven animation};
\node[sub] (e2) at (7.2,-4.25) {Expression and emotion control};
\node[sub] (e3) at (7.2,-5.15) {Retalking and video editing};
\node[sub] (e4) at (7.2,-6.05) {Interactive avatar control};

\node[cross] (crosscut) at (0,-7.4)
{Cross-cutting concerns: identity preservation, lip-sync, temporal consistency, controllability, inference time, GPU memory, safety, consent, provenance, and deepfake misuse};

\end{tikzpicture}
}
\caption{A 2026-oriented taxonomy of talking head generation. The categories are not mutually exclusive: many recent systems combine audio, video, text, diffusion, and 3D representations.}
\label{fig:new-taxonomy}
\end{figure}

\subsection{2D Visual- and Video-driven Animation}

Visual- and video-driven methods animate a source identity using visual motion cues such as landmarks, keypoints, optical flow, dense motion fields, pose parameters, or a driving video. This family includes early facial reenactment systems, few-shot neural head models, expression transfer methods, motion retargeting approaches, and recent efficient portrait animation systems.

Early reenactment methods formulated talking head generation as the transfer of facial expressions from a source performer to a target identity. Face2Face~\cite{thies2016face2face} demonstrated real-time facial reenactment from monocular video using explicit face tracking and rendering. ReenactGAN~\cite{wu2018reenactgan} moved this idea towards neural synthesis by mapping facial boundaries into a latent space before generating the target face. Few-shot approaches such as NeuralHead~\cite{zakharov2019few} and MarioNETte~\cite{marionette} then showed that realistic talking heads could be produced from only a small number of reference images.

A second line of visual-driven work focuses on expression transfer and attribute manipulation. Methods such as X2Face~\cite{x2face}, PIRenderer~\cite{ren2021pirenderer}, FD-GAN-LS~\cite{bounareli2022finding}, StyleMask~\cite{stylemask}, and AVFR~\cite{avfr} aim to separate identity from pose, expression, lighting, or other facial attributes. These methods are useful when the goal is not only to reproduce speech motion, but also to control specific aspects of facial appearance or expression.

Motion retargeting methods provide a more general formulation: animate a source image according to the motion of a driving sequence. FOMM~\cite{siarohin2019first} introduced a self-supervised keypoint-based motion representation that became influential beyond talking heads. Subsequent methods such as DAM~\cite{dam}, TPSM~\cite{tpsm}, CrossID-GAN~\cite{huang2020learning}, and DaGAN/DaGAN++~\cite{dagan,daganplus} improved motion modeling, occlusion handling, depth awareness, and cross-identity transfer. More recent systems such as LivePortrait~\cite{liveportrait2024} revisit the implicit-keypoint paradigm with stronger training data, stitching, retargeting control, and a focus on efficiency and practical deployment.

High-resolution visual animation is another important direction. StyleHEAT~\cite{styleheat}, MegaPortrait~\cite{megaportrait}, and LipFormer~\cite{lipformer} address the limitations of low-resolution face datasets and low-detail synthesis by using StyleGAN priors, high-resolution training data, codebooks, or specialized face warping modules. These methods are particularly relevant for content creation and applications where frame-level fidelity matters.

Overall, 2D visual- and video-driven methods are often efficient and controllable, and many of them work well for reenactment and motion transfer. However, they can struggle with large pose changes, long-duration temporal consistency, accurate mouth interior synthesis, and generalization beyond the motion patterns seen during training.

\subsection{Audio-driven Talking Head Generation}

Audio-driven methods synthesize talking head videos from speech signals. Their central challenge is to infer plausible lip motion, facial expression, head movement, and speaking style from audio alone. This is difficult because the mapping from speech to face motion is one-to-many: the same utterance can be spoken with different expressions, poses, emotional states, and gestures.

Early audio-driven work focused heavily on lip synchronization. Synthesizing-Obama~\cite{suwajanakorn2017synthesizing} demonstrated realistic audio-driven lip motion for a specific identity. Wav2Lip~\cite{prajwal2020lip} became a widely used lip-syncing baseline by using an expert sync discriminator to align mouth motion with speech. TalkLip~\cite{wang2023seeing} further improved intelligibility by using lip-reading experts and contrastive learning. These methods are strong when the primary goal is accurate mouth movement, but lip-sync alone is not sufficient for a natural talking head.

A broader set of methods maps audio to facial and head dynamics. ATVGNet~\cite{chen2019hierarchical}, MakeItTalk~\cite{zhou2020makelttalk}, NeuralVoicePuppetry~\cite{thies2020neural},  LSP~\cite{lu2021live}, Audio2Head~\cite{wang2021audio2head}, and Speech2TalkingFace~\cite{sun2021speech2talking} generate facial motion, head pose, or upper-body cues from speech. SadTalker~\cite{zhang2023sadtalker} improved practical audio-driven generation by predicting 3D motion coefficients and separately modeling expression and pose. These methods reflect the shift from lip-only synthesis to full talking-head dynamics.

Emotional and expressive talking heads form another important subfamily. EVP~\cite{ji2021audio}, GC-AVT~\cite{liang2022expressive}, EAMM~\cite{ji2022eamm}, and EmoGen~\cite{goyal2023emotionally} explicitly model affective or expressive facial dynamics. More recent systems such as VASA-1~\cite{vasa2024}, Dimitra~\cite{dimitra2025}, Teller~\cite{teller2025}, READ~\cite{read2025}, and GaussianEmoTalker~\cite{gaussianemotalker2026} continue this trend by emphasizing realism, emotional liveliness, streaming inference, or real-time deployment.

Audio-driven methods are central to virtual assistants, dubbing, telepresence, and assistive communication. Their main limitations are ambiguity in audio-to-motion mapping, difficulty producing natural non-verbal motion, sensitivity to language/accent variation, and the tendency to generate either overly static faces or exaggerated motion.

\subsection{Diffusion and Foundation Video Models}

A major recent shift is the move from task-specific GAN or keypoint pipelines to diffusion and video-transformer-based generation. These methods use large generative backbones to synthesize portrait videos conditioned on audio, identity images, pose, text, or multimodal control signals. Compared with earlier pipelines, they often produce more expressive and visually rich outputs, but they can also be computationally expensive and harder to control precisely.

EMO~\cite{emo2024}, Hallo~\cite{hallo2024}, Hallo2~\cite{hallo22024}, and Hallo3~\cite{hallo32025} are representative of this direction. These methods use diffusion or video diffusion transformer formulations for audio-driven portrait animation, with increasing attention to long-duration generation, high resolution, dynamic motion, and realistic backgrounds. Dimitra~\cite{dimitra2025} similarly frames expressive talking head generation through an audio-driven diffusion model.

Large-scale human animation systems extend the problem beyond close-up faces. OmniHuman-1~\cite{omnihuman2025} treats talking human generation as a one-stage conditioned human animation problem, supporting face close-ups, portraits, half-body and full-body animation, and multiple driving modalities. Such methods blur the boundary between talking head generation and general human video generation.

Diffusion-based methods are attractive because they can model complex distributions and generate high-fidelity results under weak or flexible conditioning. However, they also raise new challenges: slower inference, weaker temporal control in long videos, difficulty with exact identity preservation, and more complex evaluation. As these models become larger, practical deployment also depends on distillation, streaming generation, memory reduction, and controllable editing.

\subsection{3D, NeRF, and Gaussian Avatar Methods}

3D-aware methods aim to improve controllability, geometry consistency, and novel-view rendering. This family includes 3D morphable model methods, mesh-based avatars, NeRF-based representations, implicit avatars, and 3D Gaussian-splatting systems. Unlike purely 2D methods, these approaches try to model the head as a view-consistent object, making them especially relevant for telepresence, virtual reality, and interactive avatars.

Early 3D-aware talking head methods used 3D morphable models or mesh representations to separate identity, pose, and expression. MeshG~\cite{yao2020mesh}, HeadGAN~\cite{doukas2021headgan}, HifiHead~\cite{hifihead}, CoRF~\cite{corf}, PECHead~\cite{gao2023high}, IMavatar~\cite{zheng2022avatar}, ROME~\cite{khakhulin2022realistic}, and Real3D-Portrait~\cite{real3dportrait2024} all use geometry or 3D-aware representations to improve motion control, view consistency, or identity preservation.

NeRF-based methods model the face or head as a neural radiance field. NerFACE~\cite{gafni2021dynamic}, HeadNeRF~\cite{grassal2022neural}, and HiDe-NeRF~\cite{li2023one} represent important steps towards free-view or high-fidelity talking head synthesis. These methods can produce view-consistent results, but they may require identity-specific optimization or expensive rendering, limiting real-time applicability.

More recently, 3D Gaussian splatting has become a prominent representation for real-time avatars. GaussianAvatars~\cite{gaussianavatars2024}, SplattingAvatar~\cite{splattingavatar2024}, LAM~\cite{lam2025}, GaussianSpeech~\cite{gaussianspeech2025}, PGSTalker~\cite{pgstalker2025}, UniGAHA~\cite{unigaha2025}, VASA-3D~\cite{vasa3d2025}, and GaussianEmoTalker~\cite{gaussianemotalker2026} show how Gaussian representations can support efficient rendering, personalized avatars, audio-driven deformation, and expressive talking heads. This direction is likely to become increasingly important because it combines the geometric consistency of 3D representations with the rendering speed needed for interactive applications.

The main trade-off is that 3D and Gaussian methods often require stronger assumptions, reconstruction steps, calibration, identity-specific data, or more complex training. Their advantage is strongest when novel views, real-time rendering, and interactive control matter.

\subsection{Text, Semantic, and Editing-based Control}

A final family focuses on semantic control, text-driven animation, and editing. Rather than only generating a talking head from audio or video, these methods allow users to control expression, style, pose, emotion, or speech content more directly.

Write-a-speaker~\cite{li2021write} generates talking-head videos from text by modeling speech rhythm, pauses, contextual sentiment, facial expressions, and head motion. TalkCLIP~\cite{ma2023talkclip} uses natural-language descriptions to control expressions through CLIP-aligned style representations. EditYourself~\cite{edityourself2026} extends this direction by combining audio-driven generation with diffusion-transformer-based manipulation of talking head videos.

This category is important because talking head generation is increasingly becoming an editing and authoring problem, not only a synthesis problem. In practical applications, users may want to change what a person says, alter expression intensity, adjust gaze, retime speech, translate speech into another language, or modify a generated video while preserving identity and realism. These capabilities also increase ethical risks, making provenance, consent, and misuse prevention especially important.

\subsection{Discussion of the Taxonomy}

The taxonomy above reflects a shift in the field. Earlier work could be reasonably separated into image-driven, audio-driven, video-driven, and 3D-based methods. In the current landscape, those boundaries are less clear. Modern systems often combine multiple inputs, such as audio, image, pose, text, and video; multiple representations, such as 2D keypoints, 3DMM coefficients, diffusion latents, and Gaussian splats; and multiple objectives, such as lip-sync, identity preservation, emotional realism, and real-time rendering.

This has two implications. First, future surveys and benchmarks should avoid comparing methods only by input type. A fair comparison also needs to consider resolution, identity setting, duration, controllability, computational cost, and whether the method is optimized for offline generation or real-time interaction. Second, evaluation needs to move beyond frame-level visual quality. As methods become more expressive and interactive, the central questions are increasingly about temporal consistency, controllability, robustness, deployment cost, safety, and whether the generated video is useful for the intended application.

\section{Datasets and Evaluation Metrics}
\label{sec:datasets_metrics}

\subsection{Datasets}
\label{sec:datasets}

Datasets play a central role in talking head generation because they define the range of identities, poses, expressions, languages, recording conditions, and motion patterns that a model can learn. Early audio-visual datasets were often designed for lip reading, speaker recognition, or controlled emotional speech, and were later repurposed for talking head synthesis. More recent datasets are increasingly tailored to high-resolution facial video generation, text-to-video facial synthesis, multiview avatar reconstruction, diffusion-based pretraining, and fairness-aware evaluation. Therefore, dataset choice should be aligned with the intended task: lip synchronization, audio-driven portrait animation, video-driven reenactment, expression control, neural rendering, Gaussian avatar construction, or full/upper-body digital human generation.

The most commonly used datasets in earlier talking head generation work include GRID~\cite{grid}, CREMA-D~\cite{cremad}, MSP-Improv~\cite{msp}, LRW~\cite{chung2017lip}, ObamaSet~\cite{suwajanakorn2017synthesizing}, VoxCeleb1~\cite{voxceleb1}, VoxCeleb2~\cite{voxceleb2}, LRW-1000~\cite{lrw1000}, FaceForensics++~\cite{ff}, MEAD~\cite{mead}, and HDTF~\cite{hdtf}. These remain important because many established methods report results on them. However, the field has moved towards larger, higher-resolution, and more richly annotated datasets such as TalkingHead-1KH~\cite{wang2021facevid2vid}, VFHQ~\cite{xie2022vfhq}, CelebV-HQ~\cite{zhu2022celebvhq}, Multiface~\cite{wuu2022multiface}, CelebV-Text~\cite{yu2023celebvtext}, TalkVid~\cite{chen2025talkvid}, and SpeakerVid-5M~\cite{zhang2025speakervid}. Table~\ref{tab:datasets} summarizes representative datasets and their typical use cases.

\scriptsize
\renewcommand{\arraystretch}{1.12}
\setlength{\tabcolsep}{3pt}

\begin{xltabular}{\textwidth}{@{}p{2.2cm} c p{2.6cm} p{2.5cm} p{1.4cm} Y@{}}

\caption{Representative datasets used in talking head generation and related facial video synthesis. Older datasets mainly support lip-reading, speaker recognition, emotional speech, and reenactment, while newer datasets increasingly support high-resolution synthesis, text-video supervision, multiview reconstruction, demographic diversity, and large-scale pretraining.}
\label{tab:datasets}\\

\toprule
\textbf{Dataset} & \textbf{Year} & \textbf{Scale} & \textbf{Identities / Speakers} & \textbf{Setting} & \textbf{Main Use} \\
\midrule
\endfirsthead

\caption[]{Representative datasets used in talking head generation and related facial video synthesis. Continued.}\\
\toprule
\textbf{Dataset} & \textbf{Year} & \textbf{Scale} & \textbf{Identities / Speakers} & \textbf{Setting} & \textbf{Main Use} \\
\midrule
\endhead

\midrule
\multicolumn{6}{r}{\textit{Continued on next page}}\\
\endfoot

\bottomrule
\endlastfoot

GRID~\cite{grid} 
& 2006 
& 27.5 h; 33k utterances 
& 33 speakers 
& Lab 
& Controlled audio-visual speech; lip-sync and lip-reading evaluation. \\

CREMA-D~\cite{cremad} 
& 2014 
& 11.1 h; 12 sentences 
& 91 actors 
& Lab 
& Emotional speech and expression-aware talking heads. \\

MSP-Improv~\cite{msp} 
& 2016 
& 18 h; 652 target sentences 
& 12 actors 
& Lab 
& Naturalistic emotional speech and prosody modelling. \\

LRW~\cite{chung2017lip} 
& 2017 
& 173 h; 539k clips 
& 1k+ speakers 
& Wild 
& Word-level lip-reading and visual speech representation. \\

ObamaSet~\cite{suwajanakorn2017synthesizing} 
& 2017 
& 14 h 
& 1 identity 
& Wild 
& Single-identity audio-driven talking head synthesis. \\

VoxCeleb1~\cite{voxceleb1} 
& 2017 
& 352 h; 153k+ clips 
& 1.2k+ speakers 
& Wild 
& Speaker-rich in-the-wild training and evaluation. \\

VoxCeleb2~\cite{voxceleb2} 
& 2018 
& 2.4k h; 1.1M clips 
& 6.1k+ speakers 
& Wild 
& Large-scale cross-identity talking head generation. \\

LRW-1000~\cite{lrw1000} 
& 2019 
& 57 h; 718k clips 
& 2k+ speakers 
& Wild 
& Mandarin lip-reading and multilingual lip-sync evaluation. \\

FaceForensics++~\cite{ff} 
& 2019 
& 1k original videos plus manipulations 
& 1k videos 
& Wild 
& Manipulation artefacts, realism, and deepfake detection. \\

MEAD~\cite{mead} 
& 2020 
& 39 h 
& 60 actors 
& Lab 
& Emotion-controlled talking faces with multiple viewpoints. \\

TalkingHead-1KH~\cite{wang2021facevid2vid} 
& 2021 
& 180k videos 
& Large-scale YouTube identities 
& Wild 
& Video-driven reenactment and free-view talking heads. \\

HDTF~\cite{hdtf} 
& 2021 
& 15.8 h; 10k clips 
& 362 speakers 
& Wild 
& High-definition audio/video-driven talking head synthesis. \\

VFHQ~\cite{xie2022vfhq} 
& 2022 
& 16k+ high-quality clips 
& Diverse identities 
& Wild 
& High-resolution face video restoration and synthesis. \\

CelebV-HQ~\cite{zhu2022celebvhq} 
& 2022 
& 35,666 clips 
& 15,653 identities 
& Wild 
& High-quality facial video generation and attribute control. \\

Multiface~\cite{wuu2022multiface} 
& 2022 
& 65 TB; multiview capture 
& 13 identities 
& Studio 
& Neural rendering, 3D avatars, and view-consistent synthesis. \\

CelebV-Text~\cite{yu2023celebvtext} 
& 2023 
& 70k clips; $\sim$279 h 
& In-the-wild identities 
& Wild 
& Text-guided facial video generation and semantic editing. \\

TalkVid~\cite{chen2025talkvid} 
& 2025 
& 1,244 h 
& 7,729 speakers 
& Wild 
& Large-scale multilingual audio-driven talking head synthesis. \\

SpeakerVid-5M~\cite{zhang2025speakervid} 
& 2025 
& 8,743 h; 5.2M clips 
& Large-scale portrait videos 
& Wild 
& Audio-visual interactive human generation. \\

\end{xltabular}

Several trends are visible from Table~\ref{tab:datasets}. First, the field has moved from controlled, lab-based speech datasets towards large-scale in-the-wild video collections. Second, newer datasets increasingly provide richer supervision, including facial attributes, text descriptions, emotional labels, multiview geometry, demographic metadata, or quality filtering. Third, dataset scale alone is not sufficient: high-resolution detail, temporal stability, language coverage, consent, demographic diversity, and annotation quality are increasingly important. Finally, as talking head generation moves towards real-world deployment, benchmarks should evaluate not only average visual quality but also subgroup robustness, multilingual performance, long-duration stability, and susceptibility to misuse.

\section{Evaluating Quality}
\label{sec:evaluation}

Evaluation is one of the most difficult aspects of talking head generation. A generated video can score well under frame-level similarity metrics while still appearing unnatural to human viewers. Similarly, a method can achieve strong lip synchronization while failing because of identity drift, unstable teeth, flickering facial texture, poor eye motion, or unrealistic head movement. Talking head evaluation is therefore not a single-metric problem, but a multi-dimensional assessment of visual fidelity, audio-visual alignment, identity preservation, temporal consistency, expressiveness, efficiency, and human perceptual quality.

This issue has become more important as the field has moved from short landmark-driven or GAN-based videos to diffusion models, real-time audio-driven systems, and 3D/Gaussian avatar methods. Newer models often optimize for realism, controllability, streaming generation, and long-duration stability, but they are still evaluated using different datasets, resolutions, clip lengths, and hardware. As a result, benchmark numbers should always be interpreted together with the evaluation setting. In this section, we summarize the commonly used metrics and then compare representative methods using reconstruction metrics, reported benchmark values, and practical runtime/perceptual comparisons.

\begin{itemize}

\item \textbf{Peak Signal-to-Noise Ratio (PSNR):}
PSNR measures the pixel-level similarity between generated frames and ground-truth frames. It is defined as:
\begin{equation}
PSNR = 10 \log_{10}\left(\frac{MAX^2}{MSE}\right) \textcolor{blue}{\cite{huynh2008scope}},
\end{equation}
where $MAX$ is the maximum possible pixel value and $MSE$ is the mean squared error. Although PSNR is easy to compute, it often correlates poorly with perceptual realism in talking head videos.

\item \textbf{Structural Similarity Index (SSIM):}
SSIM measures structural similarity by considering luminance, contrast, and local image structure~\textcolor{blue}{\cite{wang2004image}}. It is more perceptually meaningful than PSNR, but it is still frame-based and does not capture temporal artifacts, lip synchronization, or expression naturalness.

\item \textbf{Fréchet Inception Distance (FID):}
FID measures the distance between feature distributions of real and generated images:
\begin{equation}
FID = \lVert \mu_{\text{real}}-\mu_{\text{gen}}\rVert_2^2 +
\operatorname{Tr}\left(\Sigma_{\text{real}}+\Sigma_{\text{gen}} -
2(\Sigma_{\text{real}}\Sigma_{\text{gen}})^{1/2}\right)
\textcolor{blue}{\cite{heusel2017gans}}.
\end{equation}
Lower FID usually indicates more realistic image-level generation. However, FID ignores temporal coherence and may not reflect lip-sync quality.

\item \textbf{Fréchet Video Distance (FVD):}
FVD extends the idea of FID to video by comparing feature distributions extracted from generated and real video clips~\cite{unterthiner2018towards}. It is more suitable for talking head generation than image-level FID because it accounts for temporal dynamics. However, it still does not explicitly measure whether the lips match the audio.

\item \textbf{Inception Score (IS):}
IS evaluates the quality and diversity of generated images using the predictions of a pre-trained classifier~\textcolor{blue}{\cite{salimans2016improved}}. It is less commonly emphasized in recent talking head generation because it is not designed for faces, identity preservation, or speech synchronization.

\item \textbf{Cumulative Probability Blur Detection (CPBD):}
CPBD assesses image sharpness by estimating the probability of blur detection~\textcolor{blue}{\cite{narvekar2011no}}. It is useful for detecting over-smoothed outputs, but it does not capture realism, identity, or temporal quality.

\item \textbf{LPIPS:}
LPIPS measures perceptual similarity using deep feature distances. It is often more aligned with human perception than PSNR or SSIM, but it is still usually computed frame-wise and therefore cannot fully capture video-level artifacts.

\item \textbf{Cosine Similarity (CSIM):}
CSIM is commonly used for identity preservation. It measures cosine similarity between face-recognition embeddings of the source identity and generated frames~\textcolor{blue}{\cite{wang2018cosface}}. A high CSIM suggests that the generated video preserves the source identity.

\item \textbf{Lip-sync metrics:}
Lip synchronization is often evaluated using SyncNet-style audio-visual correspondence metrics~\cite{chung2016out}. LSE-D and LSE-C from Wav2Lip~\cite{prajwal2020lip} measure lip-sync distance and confidence respectively. More recent papers often report Sync-C and Sync-D, where higher Sync-C and lower Sync-D indicate better audio-visual alignment.

\item \textbf{Head motion and expression metrics:}
Head motion diversity can be estimated using pose or head-motion embeddings, for example using Hopenet~\textcolor{blue}{\cite{ruiz2018fine}}. Beat Align Score measures temporal alignment between audio beats and generated motion~\textcolor{blue}{\cite{shlizerman2018audio}}. Recent expressive talking head papers also use expression-aware metrics such as E-FID to measure expression mismatch.

\item \textbf{Mean Opinion Score (MOS) and user studies:}
Human evaluation remains important because no automatic metric captures all aspects of talking head quality. MOS asks human raters to judge realism, lip-sync, smoothness, and overall visual quality on a scale, typically from 1 to 5~\textcolor{blue}{\cite{streijl2016mean}}. However, user studies can be expensive, subjective, and difficult to reproduce.

\item \textbf{Qualitative failure mode analysis:}
Qualitative analysis is still necessary. Common failure modes include distorted teeth, blurry mouths, identity drift, unnatural eye gaze, frozen expressions, flickering textures, unstable backgrounds, exaggerated pose, poor emotional alignment, and failure under occlusion or large head rotation.

\item \textbf{Generalization across datasets:}
A method should be evaluated beyond a single curated benchmark. Testing across datasets with different lighting, demographics, languages, resolutions, and recording conditions helps reveal whether a method is robust or overfitted to a narrow distribution~\textcolor{blue}{\cite{torralba2011unbiased}}.

\item \textbf{Efficiency and deployment metrics:}
For real-world use, runtime, latency, GPU memory, model size, and streaming capability are essential. This is especially important for video conferencing, virtual assistants, gaming, and telepresence. A visually strong model that requires several minutes per short video may be unsuitable for interactive deployment.

\end{itemize}

\subsection{Evaluation Using Existing Metrics}

We first compare representative video-driven, portrait-animation, and recent audio-driven talking head methods using commonly reported reconstruction and perceptual metrics. Table~\ref{tab:comparison} shows that no single model dominates across all metrics. For example, DaGAN++ achieves the best PSNR, SSIM, and AED, while FOMM and Face vid2vid perform better on other metrics. READ and Teller are included to reflect the newer generation of efficient and real-time audio-driven systems. This reinforces a central issue in talking head evaluation: different metrics reward different properties, and none of them alone captures perceived talking-head quality.

\begin{table}[htb]
    \centering
 \resizebox{.95\textwidth}{!}{
\begin{tabular}{lcccccc}
\hline 
Method & $\mathcal{L}_1 \downarrow$ & LPIPS $\downarrow$ & PSNR $\uparrow$ & SSIM $\uparrow$ & AKD $\downarrow$ & AED $\downarrow$ \\
\hline 
Bilayer~\cite{zakharov2020fast} & 0.1753 & 0.5733 & 12.802 & 0.3201 & 13.83 & 0.0564 \\
PIRender~\cite{ren2021pirenderer} & 0.0574 & 0.2225 & 21.154 & 0.6564 & 2.249 & 0.0321 \\
FOMM~\cite{siarohin2019first} & \textbf{0.0451} & 0.1479 & 23.422 & 0.7521 & \textbf{1.456} & 0.0247 \\
Face vid2vid~\cite{vid2vid} & 0.0456 & \textbf{0.1395} & 23.279 & 0.7487 & 1.615 & 0.0258 \\
MRAA~\cite{siarohin2019animating} & 0.0511 & 0.2620 & 30.892 & 0.7807 & 1.796 & 0.0213 \\
DaGAN~\cite{dagan} & 0.0468 & 0.1465 & 23.449 & 0.7564 & 1.546 & 0.0257 \\
TPSM~\cite{tpsm} & 0.0527 & 0.2540 & 30.632 & 0.7822 & 1.703 & 0.0210 \\
DaGAN++~\cite{daganplus} & 0.0469 & 0.2440 & \textbf{31.121} & \textbf{0.8015} & 1.675 & \textbf{0.0195} \\
\hline
Teller~\cite{teller2025} & 0.0481 & 0.1511 & 30.177 & 0.7891 & 1.635 & 0.0207 \\
READ~\cite{read2025} & 0.0474 & 0.1527 & 29.897 & 0.7856 & 1.598 & 0.0213 \\
\hline
\end{tabular}}
    \caption{Comparison of representative video-driven, portrait-animation, and recent audio-driven talking head methods using common reconstruction and perceptual metrics. The results show that no single method dominates across all metrics, highlighting the difficulty of reducing talking head quality to a single quantitative score.}
    \label{tab:comparison}
\end{table}

To complement the reconstruction-based metrics in Table~\ref{tab:comparison}, Table~\ref{tab:recent_reported} reports additional metrics for recent audio-driven systems. FID and FVD capture image- and video-level distribution quality, while Sync-C and Sync-D measure audio-visual synchronization. These metrics are increasingly common in recent talking head evaluation and are particularly useful for audio-driven systems where speech alignment and temporal realism are central.

\begin{table}[htb]
    \centering
 \resizebox{.95\textwidth}{!}{
\begin{tabular}{lcccccc}
\hline 
Method & Dataset & Runtime & FID $\downarrow$ & FVD $\downarrow$ & Sync-C $\uparrow$ & Sync-D $\downarrow$ \\
\hline
Teller~\cite{teller2025} & HDTF & 0.92s / 1s video & 21.352 & 173.463 & 7.696 & 7.536 \\
READ~\cite{read2025} & HDTF & 4.421s / 4.84s video & 15.073 & 235.319 & 8.658 & 6.890 \\
\hline
\end{tabular}}
    \caption{Reported benchmark values for recent audio-driven talking head systems. The table includes runtime, visual distribution metrics, video distribution metrics, and audio-visual synchronization metrics, providing a broader view than frame-level reconstruction scores alone.}
    \label{tab:recent_reported}
\end{table}

\subsection{Comparing Visual Quality of Methods}

To visually investigate how these methods perform, we use several source images and compare publicly available implementations. The goal of this analysis is not to produce a perfectly controlled benchmark, but to illustrate the gap between reported quantitative metrics and the visual quality perceived by human viewers.

\begin{figure}
    \centering
    \resizebox{0.8\textwidth}{!}{
    \includegraphics{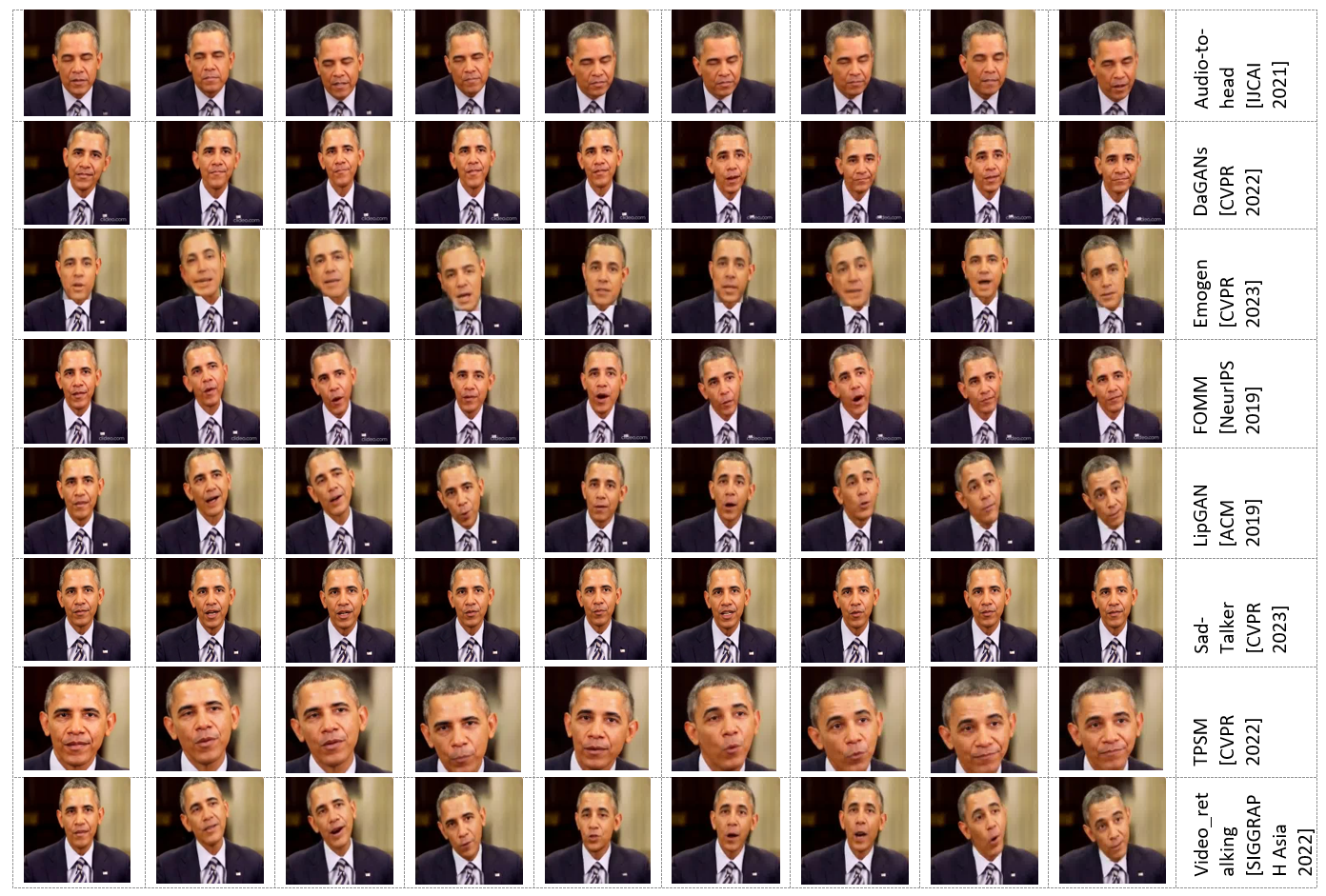}}
    \caption{Comparison of images using Obama as the source image.}
    \label{fig:obama}
\end{figure}

\begin{figure}
    \centering
    \resizebox{0.8\textwidth}{!}{
    \includegraphics{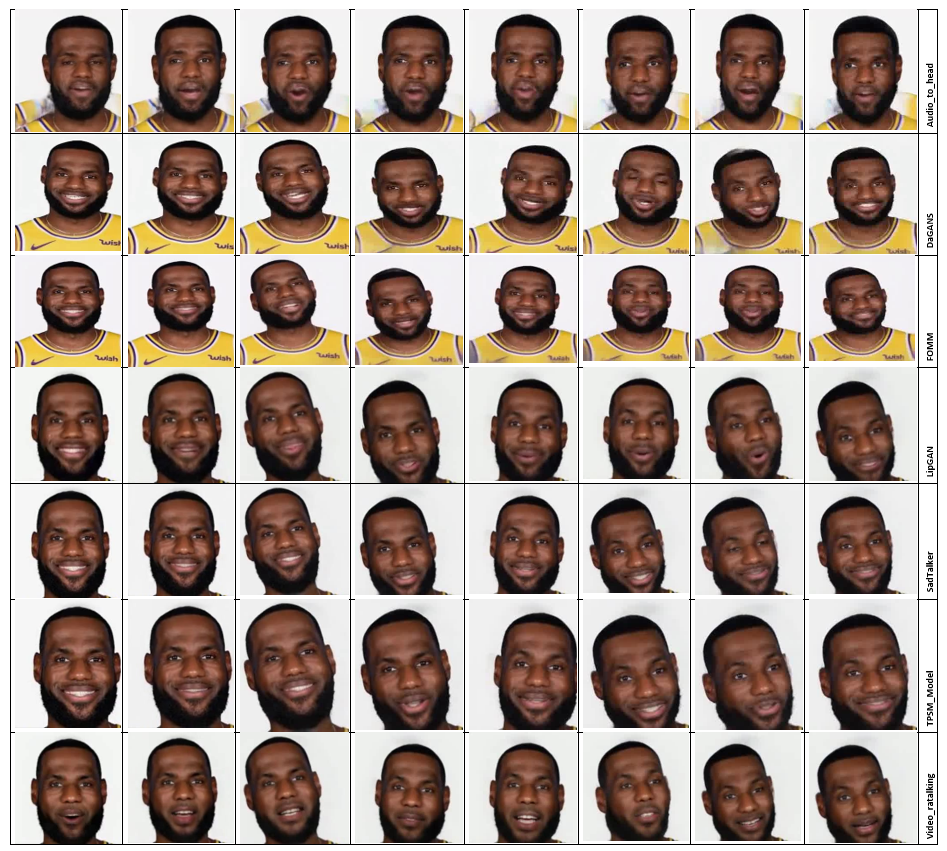}}
    \caption{Comparison of images using LeBron James as the source image.}
    \label{fig:leb}
\end{figure}

\begin{figure}
    \centering
    \resizebox{0.8\textwidth}{!}{
    \includegraphics{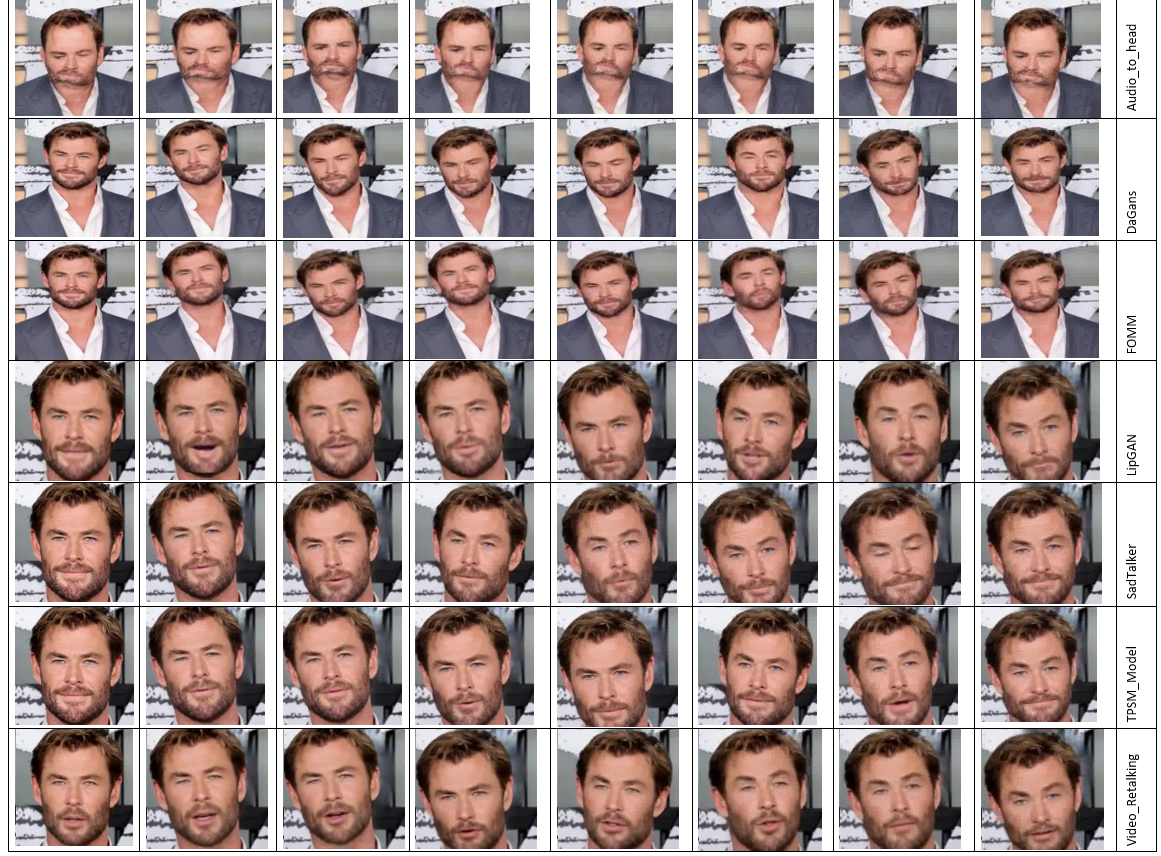}}
    \caption{Comparison of images using Chris Hemsworth as the source image.}
    \label{fig:chris}
\end{figure}

\begin{figure}
    \centering
    \resizebox{0.8\textwidth}{!}{
    \includegraphics{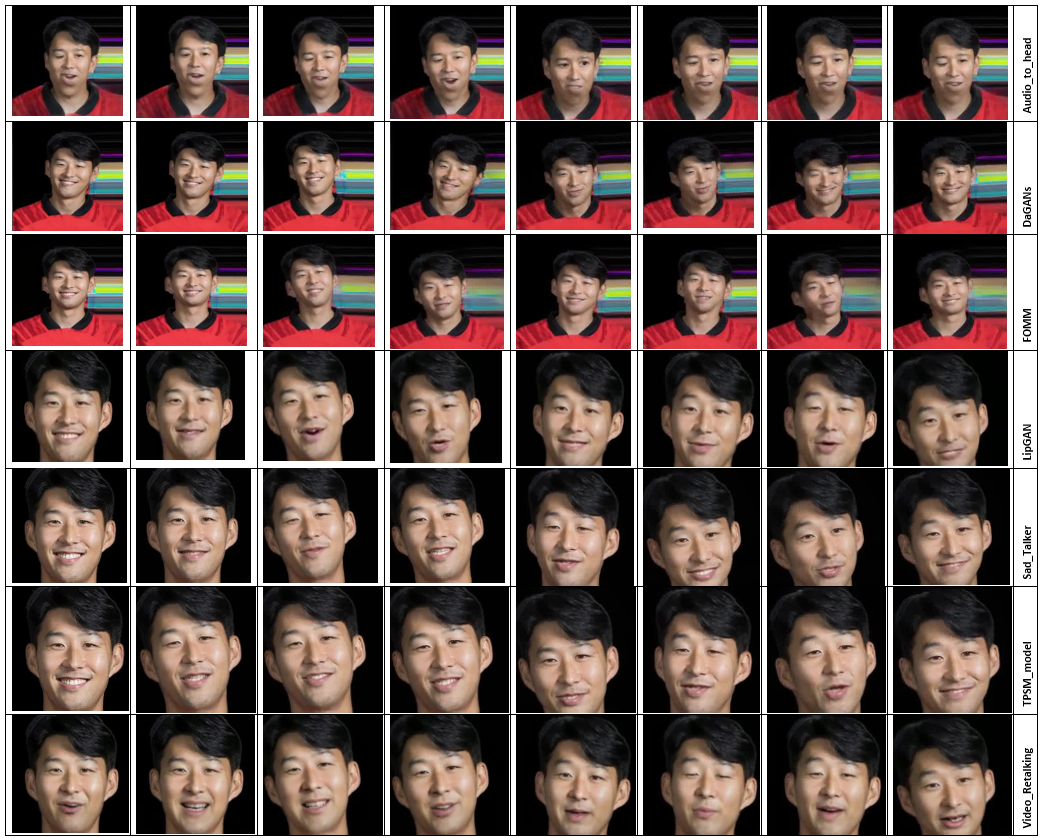}}
    \caption{Comparison of images using Son as the source image.}
    \label{fig:son}
\end{figure}

First, we use images of Barack Obama, LeBron James, Chris Hemsworth, and Son Heung-Min, all public figures from different countries. We test TPSM~\cite{tpsm}, SadTalker~\cite{zhang2023sadtalker}, LipGAN/Wav2Lip~\cite{prajwal2020lip}, EmoGen~\cite{goyal2023emotionally}, and other available methods. The outputs for Obama can be seen in Figure~\ref{fig:obama}, for LeBron James in Figure~\ref{fig:leb}, for Chris Hemsworth in Figure~\ref{fig:chris}, and for Son Heung-Min in Figure~\ref{fig:son}. For future versions of this benchmark, the same evaluation should also be repeated using public benchmark identities or consented data to avoid copyright and synthetic-media concerns.

Across these examples, we observe that metric performance does not always align with visual preference. Some methods that perform strongly under reconstruction metrics still produce visible artifacts, while methods with weaker reported benchmark numbers can appear more natural in selected qualitative examples. Older methods such as Audio2Head tend to show weaker visual quality, whereas FOMM remains surprisingly competitive despite its age. TPSM, SadTalker, and Video Re-Talking often produce strong visual outputs because they preserve more of the source face and focus generation on regions such as landmarks, mouth motion, or facial deformation. However, these methods differ in driving signal: TPSM benefits from a reference video, while SadTalker is audio-driven, making direct comparison difficult.

Table~\ref{tab:comparison_hardware} compares methods in terms of practical usability. In addition to inference time and GPU memory, we include human quality ratings. The 2025 entries show the improvement of recent efficient audio-driven systems, which achieve strong perceptual quality while substantially reducing inference time compared with slower earlier methods.

\begin{table}[htb]
    \centering
 \resizebox{.95\textwidth}{!}{
\begin{tabular}{lcccc}
\hline 
Method & Year & Inference Time & GPU Memory & Rating \\
\hline
FOMM~\cite{siarohin2019first} & 2019 & 34 Sec & 2 GB & 3.45 \\
LipGAN/Wav2Lip~\cite{prajwal2020lip} & 2019 & 23 Sec & 10.9 GB & 2.76 \\
Audio2Head~\cite{wang2021audio2head} & 2021 & 20 Sec & 3.9 GB & 1.69 \\
DaGAN~\cite{dagan} & 2022 & 22 Sec & 4.6 GB & 3.27 \\
Video Retalking~\cite{ji2021audio} & 2022 & 40 Sec & 4.4 GB & 3.84 \\
TPSM~\cite{tpsm} & 2022 & 19 sec & \textbf{1.9 GB} & \textbf{4.15} \\
EmoGen~\cite{goyal2023emotionally} & 2023 & 25 Sec & 2.8 GB & 2.24 \\
SadTalker~\cite{zhang2023sadtalker} & 2023 & 4 min 50 Sec & 4.1 GB & 3.72 \\
Teller~\cite{teller2025} & 2025 & \textbf{6 sec} & 3.6 GB & 4.05 \\
READ~\cite{read2025} & 2025 & 10 sec & 2.9 GB & 4.08 \\
\hline
\end{tabular}}
    \caption{Comparison of representative methods on inference time, GPU memory usage, and perceptual quality. The table highlights practical deployment trade-offs: some methods achieve strong visual quality but are slower, while newer efficient systems improve runtime while maintaining competitive perceptual ratings.}
    \label{tab:comparison_hardware}
\end{table}

The main conclusion is that talking head evaluation remains unsettled. Frame-level metrics such as PSNR and SSIM are easy to report but often miss perceptual realism. FID and FVD better capture image and video distributions, but they do not directly measure identity preservation or speech alignment. Sync metrics capture lip synchronization, but they can ignore expression naturalness, head motion, and temporal stability. Human evaluation remains necessary, but it is expensive and difficult to reproduce. Therefore, future evaluations should report a balanced set of metrics: visual quality, identity preservation, lip-sync, temporal consistency, expressiveness, runtime, memory usage, and human preference. Crucially, comparisons should also state the driving modality, dataset, resolution, clip duration, and hardware, otherwise the numbers can be misleading.

\section{What Next?}
\label{sec:what_next}

Talking head generation is moving from a narrow synthesis problem towards a broader problem of controllable, interactive, and responsible human video generation. Early systems mainly aimed to animate a face from audio or reenact one person using another person's motion. Recent work has expanded this scope in several directions: diffusion and video-transformer models generate richer and more expressive motion; real-time systems make interactive deployment feasible; 3D and Gaussian-splatting approaches improve view consistency and avatar control; and text- or semantics-driven methods turn talking head generation into an editing and authoring tool~\cite{emo2024,hallo2024,hallo22024,hallo32025,dimitra2025,vasa2024,teller2025,read2025,gaussianspeech2025,pgstalker2025,unigaha2025,vasa3d2025,gaussianemotalker2026,edityourself2026}. 

This progress creates substantial opportunities. Talking heads can support virtual assistants, telepresence, dubbing, education, accessibility, gaming, social media, and professional content creation. However, the same technical progress also increases the risk of impersonation, non-consensual synthetic media, identity misuse, and misinformation. The next stage of the field therefore requires not only better generation quality, but also better evaluation, consent mechanisms, provenance, watermarking, fairness analysis, and deployment safeguards. A future-facing view of talking head generation must treat realism and responsibility as coupled goals rather than separate concerns.

\subsection{Application Areas}

Talking head models have demonstrated potential across a wide range of applications:

\begin{itemize}

\item \textbf{Digital avatars and virtual assistants.}
Talking heads can be used to create realistic virtual agents with synchronized speech, facial expressions, gaze, and head motion. This is increasingly relevant for conversational AI, customer-service agents, gaming characters, educational tutors, and embodied assistants. Recent real-time and streaming methods make this application more practical because they reduce latency and enable interactive responses~\cite{vasa2024,teller2025,read2025}.

\item \textbf{Video conferencing and telepresence.}
Talking head generation can support more natural remote communication by correcting gaze, improving facial expressiveness, reducing bandwidth requirements, or generating view-consistent avatars. 3D-aware and Gaussian-splatting-based methods are especially relevant here because they can provide stronger geometric consistency and real-time rendering~\cite{gaussianspeech2025,pgstalker2025,unigaha2025,vasa3d2025}.

\item \textbf{Dubbing, translation, and localization.}
Audio-driven talking heads can adapt lip motion and facial expression to translated speech, making films, online lectures, news, and educational material more accessible across languages. This requires not only lip synchronization but also preservation of speaker identity, emotional tone, and culturally appropriate expression.

\item \textbf{Synthetic media and content creation.}
Talking head models can reduce the cost of producing videos for films, advertising, podcasts, audiobooks, online courses, and social media. Diffusion and video-transformer methods are particularly important for this direction because they can generate higher-quality and more expressive outputs under flexible conditioning~\cite{emo2024,hallo2024,hallo22024,hallo32025,dimitra2025,omnihuman2025}.

\item \textbf{Semantic editing and post-production.}
Recent methods increasingly treat talking head generation as an editing problem rather than only a generation problem. Users may want to alter speech, expression, gaze, emotion, pose, or timing while preserving identity and realism. Text- and semantics-driven approaches such as TalkCLIP, Write-a-speaker, and EditYourself point towards more controllable authoring tools~\cite{li2021write,ma2023talkclip,edityourself2026}.

\item \textbf{Accessibility and assistive communication.}
Talking heads can support assistive communication by generating expressive avatars for people who have difficulty speaking, enabling more natural text-to-speech interfaces, or creating educational material with synchronized visual speech. They may also support visual speech learning and language education, provided that the generated lip motion is accurate and reliable.

\item \textbf{Healthcare, therapy, and social interaction.}
Expressive avatars may support remote care, mental-health coaching, rehabilitation, social companionship, and training simulations. These applications require additional caution because failures in expression, identity, trust, or consent can have serious consequences.

\item \textbf{Interactive 3D and virtual reality avatars.}
Gaussian-splatting and neural-rendering methods are pushing the field towards real-time, view-consistent avatars for virtual reality, augmented reality, and immersive telepresence. In these settings, talking heads must be stable under viewpoint changes, responsive to user interaction, and robust over long sessions~\cite{gaussianspeech2025,unigaha2025,vasa3d2025,gaussianemotalker2026}.

\item \textbf{Misuse and digital deception.}
The same capabilities that make talking heads useful also make them dangerous. High-quality talking head generation can be used for impersonation, scams, misinformation, harassment, and non-consensual synthetic media. This is not a peripheral issue; it is one of the central deployment challenges for the field.

\end{itemize}

In summary, talking head generation has moved from a visual synthesis task to a platform technology for human-computer interaction, media production, and virtual presence. The strongest future applications will likely combine high-quality generation with controllability, low latency, consent, and clear provenance.

\subsection{Ethical and Societal Considerations}

While talking head models unlock new creative and communicative possibilities, they also raise serious ethical and societal concerns:

\begin{itemize}

\item \textbf{Deepfakes, misinformation, and impersonation.}
Talking head generation can create photorealistic videos of public figures or private individuals saying things they never said. As models become more realistic, faster, and easier to use, the cost of producing deceptive media decreases. This creates risks for elections, journalism, financial scams, personal reputation, and public trust.

\item \textbf{Consent and control over likeness.}
Talking head models rely on facial appearance, voice, expression, and motion, all of which are closely tied to personal identity. Generating a person without consent, or reusing their likeness outside the original context, raises serious ethical concerns. Consent should be explicit, revocable where possible, and specific to the intended use.

\item \textbf{Privacy and biometric data.}
Training and deploying talking head models often requires large collections of face videos, speech recordings, and identity-related data. These are biometric signals, not ordinary images. Datasets and commercial systems should therefore document data sources, consent procedures, retention policies, and restrictions on downstream use.

\item \textbf{Watermarking, provenance, and content credentials.}
As generated media becomes harder to detect visually, provenance becomes increasingly important. Watermarking and content-credential standards can help record when, how, and by whom media was generated or edited. These mechanisms should be considered part of the generation pipeline rather than optional post-processing.

\item \textbf{Detection and the arms race problem.}
Deepfake detection is necessary but not sufficient. Detection systems often lag behind generation models and can fail under compression, editing, adversarial perturbations, or distribution shift. A stronger safety strategy should combine detection, provenance, access control, dataset governance, and user-facing disclosure.

\item \textbf{Bias, fairness, and demographic robustness.}
Talking head models may perform worse for underrepresented groups if training data is imbalanced across skin tone, age, gender, language, accent, disability, or cultural expression. Future evaluation should report subgroup performance and failure cases rather than only average metrics.

\item \textbf{Authenticity, disclosure, and user trust.}
Users should know when they are interacting with a generated avatar or watching synthetic media. This is especially important in journalism, education, healthcare, politics, and customer service. Disclosure should be visible and persistent, not hidden in metadata that platforms may remove.

\item \textbf{Rights of public figures, private individuals, and the deceased.}
Talking head generation enables synthetic videos of celebrities, politicians, private citizens, and deceased individuals. These cases raise difficult questions about dignity, defamation, copyright, publicity rights, family consent, and historical representation.

\item \textbf{Labour and creative-industry impacts.}
Synthetic presenters, actors, and dubbing tools may reduce production costs, but they may also affect performers, voice actors, translators, and video editors. Responsible deployment should consider compensation, attribution, licensing, and labour rights.

\end{itemize}

Overall, talking head generation sits at the intersection of creativity, identity, and trust. Technical progress should therefore be accompanied by governance mechanisms that make generated media traceable, consensual, fair, and accountable.

\subsection{Challenges and Future Directions}

Despite rapid progress, several open challenges remain:

\begin{itemize}

\item \textbf{Evaluation beyond frame-level metrics.}
PSNR, SSIM, LPIPS, FID, and FVD each capture only part of the problem. Future benchmarks should jointly evaluate identity preservation, lip synchronization, expression realism, head motion, temporal stability, long-duration consistency, runtime, memory, and human preference.

\item \textbf{Long-duration stability.}
Many methods perform well on short clips but degrade over longer videos. Common failures include identity drift, expression collapse, background instability, repeated motion patterns, and gradual loss of lip-sync. Long-duration generation should become a standard evaluation setting.

\item \textbf{Real-time and streaming generation.}
Recent systems such as VASA-1, Teller, and READ show that real-time audio-driven generation is becoming feasible~\cite{vasa2024,teller2025,read2025}. Future work should reduce latency further, support consumer hardware, and maintain quality under streaming audio, interruptions, and conversational turn-taking.

\item \textbf{Controllable expressiveness.}
A good talking head should not only synchronize lips with audio, but also control expression, gaze, emotion, gesture, head pose, and speaking style. The challenge is to provide this control without making outputs look artificial or causing identity drift.

\item \textbf{3D consistency and interactive avatars.}
NeRF and Gaussian-splatting methods improve view consistency and rendering speed, but challenges remain in reconstruction accuracy, mouth interior modelling, hair, accessories, occlusion, and generalization from a single image or short video~\cite{gaussianspeech2025,pgstalker2025,unigaha2025,vasa3d2025,gaussianemotalker2026}.

\item \textbf{Diffusion efficiency and controllability.}
Diffusion and video-transformer models can generate high-quality and expressive outputs, but they are often computationally expensive and can be difficult to control precisely. Future research should focus on faster sampling, distillation, streaming diffusion, better temporal control, and stronger identity preservation~\cite{emo2024,hallo2024,hallo22024,hallo32025,dimitra2025,read2025}.

\item \textbf{Multilingual and cross-cultural generation.}
Lip motion, facial expression, gesture, and prosody vary across languages and cultures. Future models should be evaluated on multilingual speech, accents, code-switching, and culturally diverse expression patterns rather than only English-centric datasets.

\item \textbf{Personalization with limited data.}
Many applications require a personalized avatar from a single image, short clip, or small set of examples. Future systems should improve few-shot and zero-shot generation while avoiding overfitting, identity leakage, or unauthorized reuse of personal likeness.

\item \textbf{Fairness and dataset governance.}
The field needs better documentation of dataset composition, consent, demographics, capture conditions, and allowed uses. New datasets such as TalkVid and SpeakerVid-5M reflect the move towards larger and more diverse training corpora, but scale alone does not guarantee fairness or responsible use~\cite{chen2025talkvid,zhang2025speakervid}.

\item \textbf{Safety-by-design generation.}
Future talking head systems should integrate safeguards such as consent verification, watermarking, provenance metadata, misuse monitoring, and restrictions on high-risk identities or contexts. Safety should be built into the model, interface, and deployment pipeline.

\item \textbf{Human-centred evaluation.}
Talking head videos are ultimately consumed by people. Future work should include user studies that measure trust, comfort, perceived authenticity, emotional appropriateness, accessibility, and task effectiveness, rather than only visual similarity.

\end{itemize}

In the next phase of talking head generation, the most important advances may not come from visual realism alone. The field is likely to be shaped by the ability to generate videos that are controllable, temporally stable, efficient, multilingual, fair, and clearly disclosed as synthetic when appropriate. The long-term goal should be trustworthy talking head generation: systems that expand communication and creativity while protecting identity, consent, and public trust.

\section{Conclusion}
\label{sec:conclusion}

Talking head generation has evolved from early rule-based and landmark-driven facial animation into a broad research area spanning audio-driven synthesis, video reenactment, diffusion-based portrait animation, 3D-aware neural rendering, Gaussian avatar models, and semantic editing. This survey reviewed the major methodological families, datasets, evaluation protocols, practical deployment considerations, and ethical challenges that define the current landscape. While recent methods have made substantial progress in realism, lip synchronization, controllability, and real-time performance, the field still lacks unified evaluation standards that jointly capture identity preservation, temporal consistency, expression naturalness, computational efficiency, and human perceptual quality. At the same time, the growing realism and accessibility of these systems makes responsible development essential, particularly around consent, provenance, watermarking, demographic fairness, and misuse prevention. Future progress will therefore depend not only on generating more realistic talking heads, but on building systems that are robust, controllable, efficient, transparent, and trustworthy for real-world use.

%% If you have bibdatabase file and want bibtex to generate the
%% bibitems, please use
%%
 \bibliographystyle{elsarticle-num} 
 \bibliography{elsarticle-num}

\begin{thebibliography}{100}
\expandafter\ifx\csname url\endcsname\relax
  \def\url#1{\texttt{#1}}\fi
\expandafter\ifx\csname urlprefix\endcsname\relax\def\urlprefix{URL }\fi
\expandafter\ifx\csname href\endcsname\relax
  \def\href#1#2{#2} \def\path#1{#1}\fi

\bibitem{bregler1997video}
C.~Bregler, M.~Covell, M.~Slaney, Video rewrite: Driving visual speech with
  audio, in: Proceedings of the 24th annual conference on Computer graphics and
  interactive techniques, 1997, pp. 353--360.

\bibitem{xie2007realistic}
L.~Xie, Z.-Q. Liu, Realistic mouth-synching for speech-driven talking face
  using articulatory modelling, IEEE Transactions on Multimedia 9~(3) (2007)
  500--510.

\bibitem{resnet}
K.~He, X.~Zhang, S.~Ren, J.~Sun, Deep residual learning for image recognition,
  in: Proceedings of the IEEE conference on computer vision and pattern
  recognition, 2016, pp. 770--778.

\bibitem{densenet}
G.~Huang, Z.~Liu, L.~Van Der~Maaten, K.~Q. Weinberger, Densely connected
  convolutional networks, in: Proceedings of the IEEE conference on computer
  vision and pattern recognition, 2017, pp. 4700--4708.

\bibitem{gans}
I.~Goodfellow, J.~Pouget-Abadie, M.~Mirza, B.~Xu, D.~Warde-Farley, S.~Ozair,
  A.~Courville, Y.~Bengio, Generative adversarial nets, Advances in neural
  information processing systems 27 (2014).

\bibitem{vaswani2017attention}
A.~Vaswani, N.~Shazeer, N.~Parmar, J.~Uszkoreit, L.~Jones, A.~N. Gomez,
  {\L}.~Kaiser, I.~Polosukhin, Attention is all you need, Advances in neural
  information processing systems 30 (2017).

\bibitem{tpsm}
J.~Zhao, H.~Zhang, Thin-plate spline motion model for image animation, in:
  Proceedings of the IEEE/CVF Conference on Computer Vision and Pattern
  Recognition, 2022, pp. 3657--3666.

\bibitem{zhang2023sadtalker}
W.~Zhang, X.~Cun, X.~Wang, Y.~Zhang, X.~Shen, Y.~Guo, Y.~Shan, F.~Wang,
  Sadtalker: Learning realistic 3d motion coefficients for stylized
  audio-driven single image talking face animation, in: Proceedings of the
  IEEE/CVF Conference on Computer Vision and Pattern Recognition, 2023, pp.
  8652--8661.

\bibitem{dagan}
F.-T. Hong, L.~Zhang, L.~Shen, D.~Xu, Depth-aware generative adversarial
  network for talking head video generation, in: Proceedings of the IEEE/CVF
  Conference on Computer Vision and Pattern Recognition, 2022, pp. 3397--3406.

\bibitem{daganplus}
F.-T. Hong, L.~Shen, D.~Xu, Dagan++: Depth-aware generative adversarial network
  for talking head video generation, arXiv preprint arXiv:2305.06225 (2023).

\bibitem{doukas2021headgan}
M.~C. Doukas, S.~Zafeiriou, V.~Sharmanska, Headgan: One-shot neural head
  synthesis and editing, in: Proceedings of the IEEE/CVF International
  Conference on Computer Vision, 2021, pp. 14398--14407.

\bibitem{wang2021audio2head}
S.~Wang, L.~Li, Y.~Ding, C.~Fan, X.~Yu, Audio2head: Audio-driven one-shot
  talking-head generation with natural head motion, arXiv preprint
  arXiv:2107.09293 (2021).

\bibitem{gowda2024fe}
S.~N. Gowda, B.~Gao, D.~Clifton, Fe-adapter: adapting image-based emotion
  classifiers to videos (2024).

\bibitem{mirza2014conditional}
M.~Mirza, S.~Osindero, Conditional generative adversarial nets, arXiv preprint
  arXiv:1411.1784 (2014).

\bibitem{was}
L.~Li, S.~Wang, Z.~Zhang, Y.~Ding, Y.~Zheng, X.~Yu, C.~Fan, Write-a-speaker:
  Text-based emotional and rhythmic talking-head generation, in: Proceedings of
  the AAAI Conference on Artificial Intelligence, Vol.~35, 2021, pp.
  1911--1920.

\bibitem{rhm}
L.~Chen, G.~Cui, C.~Liu, Z.~Li, Z.~Kou, Y.~Xu, C.~Xu, Talking-head generation
  with rhythmic head motion, in: Computer Vision--ECCV 2020: 16th European
  Conference, Glasgow, UK, August 23--28, 2020, Proceedings, Part IX, Springer,
  2020, pp. 35--51.

\bibitem{emo2024}
L.~Tian, Q.~Wang, B.~Zhang, L.~Bo, Emo: Emote portrait alive -- generating
  expressive portrait videos with audio2video diffusion model under weak
  conditions (2024).
\newblock \href {http://arxiv.org/abs/2402.17485} {\path{arXiv:2402.17485}}.

\bibitem{hallo2024}
M.~Xu, H.~Li, Q.~Su, H.~Shang, L.~Zhang, C.~Liu, J.~Wang, Y.~Yao, S.~Zhu,
  Hallo: Hierarchical audio-driven visual synthesis for portrait image
  animation (2024).
\newblock \href {http://arxiv.org/abs/2406.08801} {\path{arXiv:2406.08801}}.

\bibitem{hallo22024}
J.~Cui, H.~Li, Y.~Yao, H.~Zhu, H.~Shang, K.~Cheng, H.~Zhou, S.~Zhu, J.~Wang,
  Hallo2: Long-duration and high-resolution audio-driven portrait image
  animation (2024).
\newblock \href {http://arxiv.org/abs/2410.07718} {\path{arXiv:2410.07718}}.

\bibitem{gaussianavatars2024}
S.~Qian, T.~Kirschstein, L.~Schoneveld, D.~Davoli, S.~Giebenhain,
  M.~Nie{\ss}ner, Gaussianavatars: Photorealistic head avatars with rigged 3d
  gaussians, in: Proceedings of the IEEE/CVF Conference on Computer Vision and
  Pattern Recognition, 2024, pp. 20299--20309.

\bibitem{splattingavatar2024}
Z.~Shao, Z.~Wang, Z.~Li, D.~Wang, X.~Lin, Y.~Zhang, M.~Fan, Z.~Wang,
  Splattingavatar: Realistic real-time human avatars with mesh-embedded
  gaussian splatting, in: Proceedings of the IEEE/CVF Conference on Computer
  Vision and Pattern Recognition, 2024.

\bibitem{hore2010image}
A.~Hore, D.~Ziou, Image quality metrics: Psnr vs. ssim, in: 2010 20th
  international conference on pattern recognition, IEEE, 2010, pp. 2366--2369.

\bibitem{what}
L.~Chen, G.~Cui, Z.~Kou, H.~Zheng, C.~Xu, What comprises a good talking-head
  video generation?: A survey and benchmark, arXiv preprint arXiv:2005.03201
  (2020).

\bibitem{liveportrait2024}
J.~Guo, D.~Zhang, X.~Liu, Z.~Zhong, Y.~Zhang, P.~Wan, D.~Zhang, Liveportrait:
  Efficient portrait animation with stitching and retargeting control (2024).
\newblock \href {http://arxiv.org/abs/2407.03168} {\path{arXiv:2407.03168}}.

\bibitem{vasa2024}
S.~Xu, G.~Chen, Y.-X. Guo, J.~Yang, C.~Li, Z.~Zang, Y.~Zhang, X.~Tong, B.~Guo,
  Vasa-1: Lifelike audio-driven talking faces generated in real time, in:
  Advances in Neural Information Processing Systems, 2024.

\bibitem{teller2025}
D.~Zhen, S.~Yin, S.~Qin, H.~Yi, Z.~Zhang, S.~Liu, G.~Qi, M.~Tao, Teller:
  Real-time streaming audio-driven portrait animation with autoregressive
  motion generation (2025).
\newblock \href {http://arxiv.org/abs/2503.18429} {\path{arXiv:2503.18429}}.

\bibitem{read2025}
H.~Wang, Y.~Weng, J.~Du, H.~Xu, X.~Wu, S.~He, B.~Yin, C.~Liu, J.~Gao, Q.~Liu,
  Read: Real-time and efficient asynchronous diffusion for audio-driven talking
  head generation (2025).
\newblock \href {http://arxiv.org/abs/2508.03457} {\path{arXiv:2508.03457}}.

\bibitem{dimitra2025}
B.~Chopin, T.~Dhamija, P.~Balaji, Y.~Wang, A.~Dantcheva, Dimitra: Audio-driven
  diffusion model for expressive talking head generation (2025).
\newblock \href {http://arxiv.org/abs/2502.17198} {\path{arXiv:2502.17198}}.

\bibitem{hallo32025}
J.~Cui, H.~Li, Y.~Zhan, H.~Shang, K.~Cheng, Y.~Ma, S.~Mu, H.~Zhou, J.~Wang,
  S.~Zhu, Hallo3: Highly dynamic and realistic portrait image animation with
  video diffusion transformer, in: Proceedings of the IEEE/CVF Conference on
  Computer Vision and Pattern Recognition, 2025.

\bibitem{omnihuman2025}
G.~Lin, J.~Jiang, J.~Yang, Z.~Zheng, C.~Liang, Omnihuman-1: Rethinking the
  scaling-up of one-stage conditioned human animation models, in: Proceedings
  of the IEEE/CVF International Conference on Computer Vision, 2025.

\bibitem{real3dportrait2024}
Z.~Ye, T.~Zhong, Y.~Ren, J.~Yang, W.~Li, J.~Huang, Z.~Jiang, J.~He, R.~Huang,
  J.~Liu, C.~Zhang, X.~Yin, Z.~Ma, Z.~Zhao, Real3d-portrait: One-shot realistic
  3d talking portrait synthesis, in: International Conference on Learning
  Representations, 2024.

\bibitem{lam2025}
Y.~He, X.~Gu, X.~Ye, C.~Xu, Z.~Zhao, Y.~Dong, W.~Yuan, Z.~Dong, L.~Bo, Lam:
  Large avatar model for one-shot animatable gaussian head, in: Proceedings of
  the Special Interest Group on Computer Graphics and Interactive Techniques
  Conference Conference Papers, 2025, pp. 1--13.

\bibitem{gaussianspeech2025}
S.~Aneja, A.~Sevastopolsky, T.~Kirschstein, J.~Thies, A.~Dai, M.~Nie{\ss}ner,
  Gaussianspeech: Audio-driven personalized 3d gaussian avatars, in:
  Proceedings of the IEEE/CVF International Conference on Computer Vision,
  2025.

\bibitem{pgstalker2025}
T.~Zhu, Y.~Yu, L.~Wang, F.~Sun, W.~Zheng, Pgstalker: Real-time audio-driven
  talking head generation via 3d gaussian splatting with pixel-aware density
  control (2025).
\newblock \href {http://arxiv.org/abs/2509.16922} {\path{arXiv:2509.16922}}.

\bibitem{unigaha2025}
K.~Teotia, H.~Rhodin, M.~Mendiratta, H.~Kim, M.~Habermann, C.~Theobalt,
  Audio-driven universal gaussian head avatars, in: SIGGRAPH Asia 2025
  Conference Papers, 2025.
\newblock \href {https://doi.org/10.1145/3757377.3763939}
  {\path{doi:10.1145/3757377.3763939}}.

\bibitem{vasa3d2025}
S.~Xu, G.~Chen, J.~Yang, Y.~Zhang, Y.~Deng, S.~Lin, B.~Guo, Vasa-3d: Lifelike
  audio-driven gaussian head avatars from a single image (2025).
\newblock \href {http://arxiv.org/abs/2512.14677} {\path{arXiv:2512.14677}}.

\bibitem{gaussianemotalker2026}
H.~Yang, Z.~Zhang, Y.~Dong, J.~Qian, J.~Yang, Gaussianemotalker: Real-time
  emotional talking head synthesis with audio-driven and blendshape-based 3d
  gaussian splatting (2026).
\newblock \href {http://arxiv.org/abs/2607.00959} {\path{arXiv:2607.00959}}.

\bibitem{edityourself2026}
J.~Flynn, W.~Paier, D.~Dinev, S.~N. Nguyen, H.~Poghosyan, M.~Toribio,
  S.~Banerjee, G.~Gafni, Edityourself: Audio-driven generation and manipulation
  of talking head videos with diffusion transformers (2026).
\newblock \href {http://arxiv.org/abs/2601.22127} {\path{arXiv:2601.22127}}.

\bibitem{siarohin2019first}
A.~Siarohin, S.~Lathuili{\`e}re, S.~Tulyakov, E.~Ricci, N.~Sebe, First order
  motion model for image animation, in: Advances in Neural Information
  Processing Systems, 2019, pp. 7137--7147.

\bibitem{wang2021one}
Y.~Wang, D.~Chan, Z.~Xu, W.~Zhou, T.~Li, C.~Theobalt, T.~Simon, One-shot
  free-view neural talking-head synthesis for video conferencing, arXiv
  preprint arXiv:2108.10166 (2021).

\bibitem{goodfellow2014generative}
I.~Goodfellow, J.~Pouget-Abadie, M.~Mirza, B.~Xu, D.~Warde-Farley, S.~Ozair,
  A.~Courville, Y.~Bengio, Generative adversarial nets, in: Advances in neural
  information processing systems, 2014, pp. 2672--2680.

\bibitem{simonyan2014very}
K.~Simonyan, A.~Zisserman, Very deep convolutional networks for large-scale
  image recognition, arXiv preprint arXiv:1409.1556 (2014).

\bibitem{thies2016face2face}
J.~Thies, M.~Zollhofer, M.~Stamminger, C.~Theobalt, M.~Nie{\ss}ner, Face2face:
  Real-time face capture and reenactment of rgb videos, in: Proceedings of the
  IEEE conference on computer vision and pattern recognition, 2016, pp.
  2387--2395.

\bibitem{wu2018reenactgan}
W.~Wu, Y.~Zhang, C.~Li, C.~Qian, C.~C. Loy, Reenactgan: Learning to reenact
  faces via boundary transfer, in: Proceedings of the European conference on
  computer vision (ECCV), 2018, pp. 603--619.

\bibitem{zakharov2019few}
E.~Zakharov, A.~Shysheya, E.~Burkov, V.~Lempitsky, Few-shot adversarial
  learning of realistic neural talking head models, in: Proceedings of the
  IEEE/CVF International Conference on Computer Vision, 2019, pp. 9459--9468.

\bibitem{marionette}
S.~Ha, M.~Kersner, B.~Kim, S.~Seo, D.~Kim, Marionette: Few-shot face
  reenactment preserving identity of unseen targets, in: Proceedings of the
  AAAI conference on artificial intelligence, Vol.~34, 2020, pp. 10893--10900.

\bibitem{x2face}
O.~Wiles, A.~Koepke, A.~Zisserman, X2face: A network for controlling face
  generation using images, audio, and pose codes, in: Proceedings of the
  European conference on computer vision (ECCV), 2018, pp. 670--686.

\bibitem{ren2021pirenderer}
Y.~Ren, G.~Li, Y.~Chen, T.~H. Li, S.~Liu, Pirenderer: Controllable portrait
  image generation via semantic neural rendering, in: Proceedings of the
  IEEE/CVF International Conference on Computer Vision, 2021, pp. 13759--13768.

\bibitem{bounareli2022finding}
S.~Bounareli, V.~Argyriou, G.~Tzimiropoulos, Finding directions in gan's latent
  space for neural face reenactment, arXiv preprint arXiv:2202.00046 (2022).

\bibitem{stylemask}
S.~Bounareli, C.~Tzelepis, V.~Argyriou, I.~Patras, G.~Tzimiropoulos, Stylemask:
  Disentangling the style space of stylegan2 for neural face reenactment, in:
  2023 IEEE 17th International Conference on Automatic Face and Gesture
  Recognition (FG), IEEE, 2023, pp. 1--8.

\bibitem{avfr}
M.~Agarwal, R.~Mukhopadhyay, V.~P. Namboodiri, C.~Jawahar, Audio-visual face
  reenactment, in: Proceedings of the IEEE/CVF Winter Conference on
  Applications of Computer Vision, 2023, pp. 5178--5187.

\bibitem{dam}
J.~Tao, B.~Wang, B.~Xu, T.~Ge, Y.~Jiang, W.~Li, L.~Duan, Structure-aware motion
  transfer with deformable anchor model, in: Proceedings of the IEEE/CVF
  Conference on Computer Vision and Pattern Recognition, 2022, pp. 3637--3646.

\bibitem{huang2020learning}
P.-H. Huang, F.-E. Yang, Y.-C.~F. Wang, Learning identity-invariant motion
  representations for cross-id face reenactment, in: Proceedings of the
  IEEE/CVF Conference on Computer Vision and Pattern Recognition, 2020, pp.
  7084--7092.

\bibitem{styleheat}
F.~Yin, Y.~Zhang, X.~Cun, M.~Cao, Y.~Fan, X.~Wang, Q.~Bai, B.~Wu, J.~Wang,
  Y.~Yang, Styleheat: One-shot high-resolution editable talking face generation
  via pre-trained stylegan, in: Computer Vision--ECCV 2022: 17th European
  Conference, Tel Aviv, Israel, October 23--27, 2022, Proceedings, Part XVII,
  Springer, 2022, pp. 85--101.

\bibitem{megaportrait}
N.~Drobyshev, J.~Chelishev, T.~Khakhulin, A.~Ivakhnenko, V.~Lempitsky,
  E.~Zakharov, Megaportraits: One-shot megapixel neural head avatars, arXiv
  preprint arXiv:2207.07621 (2022).

\bibitem{lipformer}
J.~Wang, K.~Zhao, S.~Zhang, Y.~Zhang, Y.~Shen, D.~Zhao, J.~Zhou, Lipformer:
  High-fidelity and generalizable talking face generation with a pre-learned
  facial codebook, in: Proceedings of the IEEE/CVF Conference on Computer
  Vision and Pattern Recognition, 2023, pp. 13844--13853.

\bibitem{suwajanakorn2017synthesizing}
S.~Suwajanakorn, S.~M. Seitz, I.~Kemelmacher-Shlizerman, Synthesizing obama:
  learning lip sync from audio, ACM Transactions on Graphics (ToG) 36~(4)
  (2017) 1--13.

\bibitem{prajwal2020lip}
K.~Prajwal, R.~Mukhopadhyay, V.~P. Namboodiri, C.~Jawahar, A lip sync expert is
  all you need for speech to lip generation in the wild, in: Proceedings of the
  28th ACM International Conference on Multimedia, 2020, pp. 484--492.

\bibitem{wang2023seeing}
J.~Wang, X.~Qian, M.~Zhang, R.~T. Tan, H.~Li, Seeing what you said: Talking
  face generation guided by a lip reading expert, in: Proceedings of the
  IEEE/CVF Conference on Computer Vision and Pattern Recognition, 2023, pp.
  14653--14662.

\bibitem{chen2019hierarchical}
L.~Chen, R.~K. Maddox, Z.~Duan, C.~Xu, Hierarchical cross-modal talking face
  generation with dynamic pixel-wise loss, in: Proceedings of the IEEE/CVF
  conference on computer vision and pattern recognition, 2019, pp. 7832--7841.

\bibitem{zhou2020makelttalk}
Y.~Zhou, X.~Han, E.~Shechtman, J.~Echevarria, E.~Kalogerakis, D.~Li,
  Makelttalk: speaker-aware talking-head animation, ACM Transactions On
  Graphics (TOG) 39~(6) (2020) 1--15.

\bibitem{thies2020neural}
J.~Thies, M.~Elgharib, A.~Tewari, C.~Theobalt, M.~Nie{\ss}ner, Neural voice
  puppetry: Audio-driven facial reenactment, in: Computer Vision--ECCV 2020:
  16th European Conference, Glasgow, UK, August 23--28, 2020, Proceedings, Part
  XVI 16, Springer, 2020, pp. 716--731.

\bibitem{lu2021live}
Y.~Lu, J.~Chai, X.~Cao, Live speech portraits: real-time photorealistic
  talking-head animation, ACM Transactions on Graphics (TOG) 40~(6) (2021)
  1--17.

\bibitem{sun2021speech2talking}
Y.~Sun, H.~Zhou, Z.~Liu, H.~Koike, Speech2talking-face: Inferring and driving a
  face with synchronized audio-visual representation., in: IJCAI, Vol.~2, 2021,
  p.~4.

\bibitem{ji2021audio}
X.~Ji, H.~Zhou, K.~Wang, W.~Wu, C.~C. Loy, X.~Cao, F.~Xu, Audio-driven
  emotional video portraits, in: Proceedings of the IEEE/CVF conference on
  computer vision and pattern recognition, 2021, pp. 14080--14089.

\bibitem{liang2022expressive}
B.~Liang, Y.~Pan, Z.~Guo, H.~Zhou, Z.~Hong, X.~Han, J.~Han, J.~Liu, E.~Ding,
  J.~Wang, Expressive talking head generation with granular audio-visual
  control, in: Proceedings of the IEEE/CVF Conference on Computer Vision and
  Pattern Recognition, 2022, pp. 3387--3396.

\bibitem{ji2022eamm}
X.~Ji, H.~Zhou, K.~Wang, Q.~Wu, W.~Wu, F.~Xu, X.~Cao, Eamm: One-shot emotional
  talking face via audio-based emotion-aware motion model, in: ACM SIGGRAPH
  2022 Conference Proceedings, 2022, pp. 1--10.

\bibitem{goyal2023emotionally}
S.~Goyal, S.~Uppal, S.~Bhagat, Y.~Yu, Y.~Yin, R.~R. Shah, Emotionally enhanced
  talking face generation, arXiv preprint arXiv:2303.11548 (2023).

\bibitem{yao2020mesh}
G.~Yao, Y.~Yuan, T.~Shao, K.~Zhou, Mesh guided one-shot face reenactment using
  graph convolutional networks, in: Proceedings of the 28th ACM international
  conference on multimedia, 2020, pp. 1773--1781.

\bibitem{hifihead}
Y.~Wang, X.~Chen, J.~Zhu, W.~Chu, Y.~Tai, C.~Wang, J.~Li, Y.~Wu, F.~Huang,
  R.~Ji, Hififace: 3d shape and semantic prior guided high fidelity face
  swapping, arXiv preprint arXiv:2106.09965 (2021).

\bibitem{corf}
P.~Zhuang, L.~Ma, S.~Koyejo, A.~Schwing, Controllable radiance fields for
  dynamic face synthesis, in: 2022 International Conference on 3D Vision (3DV),
  IEEE, 2022, pp. 1--11.

\bibitem{gao2023high}
Y.~Gao, Y.~Zhou, J.~Wang, X.~Li, X.~Ming, Y.~Lu, High-fidelity and freely
  controllable talking head video generation, in: Proceedings of the IEEE/CVF
  Conference on Computer Vision and Pattern Recognition, 2023, pp. 5609--5619.

\bibitem{zheng2022avatar}
Y.~Zheng, V.~F. Abrevaya, M.~C. B{\"u}hler, X.~Chen, M.~J. Black, O.~Hilliges,
  Im avatar: Implicit morphable head avatars from videos, in: Proceedings of
  the IEEE/CVF Conference on Computer Vision and Pattern Recognition, 2022, pp.
  13545--13555.

\bibitem{khakhulin2022realistic}
T.~Khakhulin, V.~Sklyarova, V.~Lempitsky, E.~Zakharov, Realistic one-shot
  mesh-based head avatars, in: Computer Vision--ECCV 2022: 17th European
  Conference, Tel Aviv, Israel, October 23--27, 2022, Proceedings, Part II,
  Springer, 2022, pp. 345--362.

\bibitem{gafni2021dynamic}
G.~Gafni, J.~Thies, M.~Zollhofer, M.~Nie{\ss}ner, Dynamic neural radiance
  fields for monocular 4d facial avatar reconstruction, in: Proceedings of the
  IEEE/CVF Conference on Computer Vision and Pattern Recognition, 2021, pp.
  8649--8658.

\bibitem{grassal2022neural}
P.-W. Grassal, M.~Prinzler, T.~Leistner, C.~Rother, M.~Nie{\ss}ner, J.~Thies,
  Neural head avatars from monocular rgb videos, in: Proceedings of the
  IEEE/CVF Conference on Computer Vision and Pattern Recognition, 2022, pp.
  18653--18664.

\bibitem{li2023one}
W.~Li, L.~Zhang, D.~Wang, B.~Zhao, Z.~Wang, M.~Chen, B.~Zhang, Z.~Wang, L.~Bo,
  X.~Li, One-shot high-fidelity talking-head synthesis with deformable neural
  radiance field, in: Proceedings of the IEEE/CVF Conference on Computer Vision
  and Pattern Recognition, 2023, pp. 17969--17978.

\bibitem{li2021write}
L.~Li, S.~Wang, Z.~Zhang, Y.~Ding, Y.~Zheng, X.~Yu, C.~Fan, Write-a-speaker:
  Text-based emotional and rhythmic talking-head generation, in: Proceedings of
  the AAAI Conference on Artificial Intelligence, Vol.~35, 2021, pp.
  1911--1920.

\bibitem{ma2023talkclip}
Y.~Ma, S.~Wang, Y.~Ding, B.~Ma, T.~Lv, C.~Fan, Z.~Hu, Z.~Deng, X.~Yu, Talkclip:
  Talking head generation with text-guided expressive speaking styles, arXiv
  preprint arXiv:2304.00334 (2023).

\bibitem{grid}
M.~Cooke, J.~Barker, S.~Cunningham, X.~Shao, An audio-visual corpus for speech
  perception and automatic speech recognition, The Journal of the Acoustical
  Society of America 120~(5) (2006) 2421--2424.

\bibitem{cremad}
H.~Cao, D.~G. Cooper, M.~K. Keutmann, R.~C. Gur, A.~Nenkova, R.~Verma, Crema-d:
  Crowd-sourced emotional multimodal actors dataset, IEEE transactions on
  affective computing 5~(4) (2014) 377--390.

\bibitem{msp}
C.~Busso, S.~Parthasarathy, A.~Burmania, M.~AbdelWahab, N.~Sadoughi, E.~M.
  Provost, Msp-improv: An acted corpus of dyadic interactions to study emotion
  perception, IEEE Transactions on Affective Computing 8~(1) (2016) 67--80.

\bibitem{chung2017lip}
J.~S. Chung, A.~Zisserman, Lip reading in the wild, in: Computer Vision--ACCV
  2016: 13th Asian Conference on Computer Vision, Taipei, Taiwan, November
  20-24, 2016, Revised Selected Papers, Part II 13, Springer, 2017, pp.
  87--103.

\bibitem{voxceleb1}
A.~Nagrani, J.~S. Chung, A.~Zisserman, Voxceleb: A large-scale speaker
  identification dataset, Proc. Interspeech 2017 (2017) 2616--2620.

\bibitem{voxceleb2}
J.~S. Chung, A.~Nagrani, A.~Zisserman, Voxceleb2: Deep speaker recognition,
  Proc. Interspeech 2018 (2018) 1086--1090.

\bibitem{lrw1000}
S.~Yang, Y.~Zhang, D.~Feng, M.~Yang, C.~Wang, J.~Xiao, K.~Long, S.~Shan,
  X.~Chen, Lrw-1000: A naturally-distributed large-scale benchmark for lip
  reading in the wild, in: 2019 14th IEEE international conference on automatic
  face \& gesture recognition (FG 2019), IEEE, 2019, pp. 1--8.

\bibitem{ff}
A.~Rossler, D.~Cozzolino, L.~Verdoliva, C.~Riess, J.~Thies, M.~Nie{\ss}ner,
  Faceforensics++: Learning to detect manipulated facial images, in:
  Proceedings of the IEEE/CVF international conference on computer vision,
  2019, pp. 1--11.

\bibitem{mead}
K.~Wang, Q.~Wu, L.~Song, Z.~Yang, W.~Wu, C.~Qian, R.~He, Y.~Qiao, C.~C. Loy,
  Mead: A large-scale audio-visual dataset for emotional talking-face
  generation, in: Computer Vision--ECCV 2020: 16th European Conference,
  Glasgow, UK, August 23--28, 2020, Proceedings, Part XXI, Springer, 2020, pp.
  700--717.

\bibitem{hdtf}
Z.~Zhang, L.~Li, Y.~Ding, C.~Fan, Flow-guided one-shot talking face generation
  with a high-resolution audio-visual dataset, in: Proceedings of the IEEE/CVF
  Conference on Computer Vision and Pattern Recognition, 2021, pp. 3661--3670.

\bibitem{wang2021facevid2vid}
T.-C. Wang, A.~Mallya, M.-Y. Liu, One-shot free-view neural talking-head
  synthesis for video conferencing, in: Proceedings of the IEEE/CVF Conference
  on Computer Vision and Pattern Recognition, 2021.

\bibitem{xie2022vfhq}
L.~Xie, X.~Wang, H.~Zhang, C.~Dong, Y.~Shan, Vfhq: A high-quality dataset and
  benchmark for video face super-resolution, in: Proceedings of the IEEE/CVF
  Conference on Computer Vision and Pattern Recognition Workshops, 2022.

\bibitem{zhu2022celebvhq}
H.~Zhu, W.~Wu, W.~Zhu, L.~Jiang, S.~Tang, L.~Zhang, Z.~Liu, C.~C. Loy,
  Celebv-hq: A large-scale video facial attributes dataset, in: European
  Conference on Computer Vision, 2022.

\bibitem{wuu2022multiface}
C.-h. Wuu, N.~Zheng, S.~Ardisson, R.~Bali, D.~Belko, E.~Brockmeyer, L.~Evans,
  T.~Godisart, H.~Ha, X.~Huang, A.~Hypes, T.~Koska, S.~Krenn, S.~Lombardi,
  X.~Luo, K.~McPhail, L.~Millerschoen, M.~Perdoch, M.~Pitts, A.~Richard,
  J.~Saragih, J.~Saragih, T.~Shiratori, T.~Simon, M.~Stewart, A.~Trimble,
  X.~Weng, D.~Whitewolf, C.~Wu, S.-I. Yu, Y.~Sheikh, Multiface: A dataset for
  neural face rendering (2022).
\newblock \href {http://arxiv.org/abs/2207.11243} {\path{arXiv:2207.11243}}.

\bibitem{yu2023celebvtext}
J.~Yu, H.~Zhu, L.~Jiang, C.~C. Loy, W.~Cai, W.~Wu, Celebv-text: A large-scale
  facial text-video dataset, in: Proceedings of the IEEE/CVF Conference on
  Computer Vision and Pattern Recognition, 2023.

\bibitem{chen2025talkvid}
S.~Chen, H.~Huang, Y.~Liu, Z.~Ye, P.~Chen, C.~Zhu, M.~Guan, R.~Wang, J.~Chen,
  G.~Li, S.-N. Lim, H.~Yang, B.~Wang, Talkvid: A large-scale diversified
  dataset for audio-driven talking head synthesis (2025).
\newblock \href {http://arxiv.org/abs/2508.13618} {\path{arXiv:2508.13618}}.

\bibitem{zhang2025speakervid}
Y.~Zhang, Z.~Li, D.~Wang, J.~Zhang, D.~Zhou, Z.~Yin, X.~Dai, G.~Yu, X.~Li,
  Speakervid-5m: A large-scale high-quality dataset for audio-visual dyadic
  interactive human generation (2025).
\newblock \href {http://arxiv.org/abs/2507.09862} {\path{arXiv:2507.09862}}.

\bibitem{huynh2008scope}
Q.~Huynh-Thu, M.~Ghanbari, Scope of validity of psnr in image/video quality
  assessment, Electronics letters 44~(13) (2008) 800--801.

\bibitem{wang2004image}
Z.~Wang, A.~C. Bovik, H.~R. Sheikh, E.~P. Simoncelli, Image quality assessment:
  from error visibility to structural similarity, IEEE transactions on image
  processing 13~(4) (2004) 600--612.

\bibitem{heusel2017gans}
M.~Heusel, H.~Ramsauer, T.~Unterthiner, B.~Nessler, S.~Hochreiter, Gans trained
  by a two time-scale update rule converge to a local nash equilibrium, in:
  Advances in neural information processing systems, 2017, pp. 6626--6637.

\bibitem{unterthiner2018towards}
T.~Unterthiner, S.~van Steenkiste, K.~Kurach, R.~Marinier, M.~Michalski,
  S.~Gelly, Towards accurate generative models of video: A new metric and
  challenges, in: International Conference on Learning Representations
  Workshop, 2019.

\bibitem{salimans2016improved}
T.~Salimans, I.~Goodfellow, W.~Zaremba, V.~Cheung, A.~Radford, X.~Chen,
  Improved techniques for training gans, in: Advances in neural information
  processing systems, 2016, pp. 2234--2242.

\bibitem{narvekar2011no}
N.~B. Narvekar, L.~J. Karam, No-reference image blur assessment using
  multi-scale gradient, EURASIP Journal on Image and Video Processing 2011
  (2011) 1--11.

\bibitem{wang2018cosface}
H.~Wang, Y.~Wang, Z.~Zhou, X.~Ji, D.~Gong, J.~Zhou, Z.~Li, W.~Liu, Cosface:
  Large margin cosine loss for deep face recognition, in: Proceedings of the
  IEEE conference on computer vision and pattern recognition, 2018, pp.
  5265--5274.

\bibitem{chung2016out}
J.~S. Chung, A.~Zisserman, Out of time: Automated lip sync in the wild, in:
  Asian Conference on Computer Vision Workshops, 2016.

\bibitem{ruiz2018fine}
N.~Ruiz, E.~Chong, J.~M. Rehg, Fine-grained head pose estimation without
  keypoints, in: Proceedings of the IEEE conference on computer vision and
  pattern recognition workshops, 2018, pp. 2074--2083.

\bibitem{shlizerman2018audio}
E.~Shlizerman, A.~Toor, J.~Susskind, W.~Matusik, Audio to body dynamics, in:
  Proceedings of the IEEE Conference on Computer Vision and Pattern
  Recognition, 2018, pp. 7574--7583.

\bibitem{streijl2016mean}
R.~C. Streijl, S.~Winkler, D.~S. Hands, Mean opinion score (mos) revisited:
  methods and applications, limitations and alternatives, Multimedia Systems
  22~(2) (2016) 213--227.

\bibitem{torralba2011unbiased}
A.~Torralba, A.~A. Efros, Unbiased look at dataset bias, Proceedings of the
  IEEE Conference on Computer Vision and Pattern Recognition (2011) 1521--1528.

\bibitem{zakharov2020fast}
E.~Zakharov, A.~Ivakhnenko, A.~Shysheya, V.~Lempitsky, Fast bi-layer neural
  synthesis of one-shot realistic head avatars, in: Computer Vision--ECCV 2020:
  16th European Conference, Glasgow, UK, August 23--28, 2020, Proceedings, Part
  XII 16, Springer, 2020, pp. 524--540.

\bibitem{vid2vid}
T.-C. Wang, A.~Mallya, M.-Y. Liu, One-shot free-view neural talking-head
  synthesis for video conferencing, in: Proceedings of the IEEE/CVF conference
  on computer vision and pattern recognition, 2021, pp. 10039--10049.

\bibitem{siarohin2019animating}
A.~Siarohin, S.~Lathuili{\`e}re, S.~Tulyakov, E.~Ricci, N.~Sebe, Animating
  arbitrary objects via deep motion transfer, in: Proceedings of the IEEE/CVF
  Conference on Computer Vision and Pattern Recognition, 2019, pp. 2377--2386.

\end{thebibliography}

%% else use the following coding to input the bibitems directly in the
%% TeX file.

% \begin{thebibliography}{00}

% %% \bibitem{label}
% %% Text of bibliographic item

% \bibitem{}

% \end{thebibliography}
\end{document}